\documentclass[10pt,twocolumn,letterpaper]{article}

\usepackage[pagenumbers]{cvpr} % To force page numbers, e.g. for an arXiv version

\usepackage{graphicx}
\usepackage{amsmath}
\usepackage{amssymb}
\usepackage{booktabs}

\usepackage{bm}
\usepackage{siunitx}
\usepackage{stfloats}
\usepackage{multirow}
\usepackage{pifont}
\usepackage[utf8x]{inputenc} 
\newcommand{\xmark}{\ding{55}}%
\usepackage[dvipsnames, table]{xcolor}
\usepackage{makecell}

\usepackage[pagebackref,breaklinks,colorlinks]{hyperref}

\usepackage[capitalize]{cleveref}
\crefname{section}{Sec.}{Secs.}
\Crefname{section}{Section}{Sections}
\Crefname{table}{Table}{Tables}
\crefname{table}{Tab.}{Tabs.}

\def\cvprPaperID{****} % *** Enter the CVPR Paper ID here
\def\confName{CVPR}
\def\confYear{2023}

\usepackage{hhline}

\makeatletter
\newcommand\smallfootnote{\@setfontsize\smallfootnote{7.5}{8.5}}
\makeatother

\begin{document}

\title{Self-Supervised Learning based on Heat Equation}
\author{Yinpeng Chen\textsuperscript{1} \qquad Xiyang Dai\textsuperscript{1} \qquad Dongdong Chen\textsuperscript{1} \qquad Mengchen Liu\textsuperscript{1} \\
Lu Yuan\textsuperscript{1} \qquad Zicheng Liu\textsuperscript{1} \qquad Youzuo Lin\textsuperscript{2} \\
\\
\qquad \qquad \qquad \qquad \textsuperscript{1} Microsoft \qquad \qquad \qquad \qquad \qquad \textsuperscript{2} Los Alamos National Laboratory \\
{\tt\small \{yiche,xidai,dochen,mengcliu,luyuan,zliu\}@microsoft.com \qquad \qquad ylin@lanl.gov \qquad \qquad \qquad}
%\and
}

\maketitle

%%%%%%%%% ABSTRACT
\begin{abstract}
    This paper presents a new perspective of self-supervised learning based on extending heat equation into high dimensional feature space. In particular, we remove time dependence by steady-state condition, and extend the remaining 2D Laplacian from $x$--$y$ isotropic to linear correlated. Furthermore, we simplify it by splitting $x$ and $y$ axes as two first-order linear differential equations. Such simplification explicitly models the spatial invariance along horizontal and vertical directions separately, supporting prediction across image blocks. This introduces a very simple masked image modeling (MIM) method, named QB-Heat.
    
    QB-Heat leaves a single block with size of quarter image unmasked and extrapolates other three masked quarters linearly. It brings MIM to CNNs without bells and whistles, and even works well for pre-training light-weight networks that are suitable for both image classification and object detection without fine-tuning. Compared with MoCo-v2 on pre-training a Mobile-Former with 5.8M parameters and 285M FLOPs, QB-Heat is on par in linear probing on ImageNet, but clearly outperforms in non-linear probing that adds a transformer block before linear classifier (65.6\% vs. 52.9\%). When transferring to object detection with frozen backbone, QB-Heat outperforms MoCo-v2 and supervised pre-training on ImageNet by 7.9 and 4.5 AP respectively.
    
    This work provides an insightful hypothesis on the invariance within visual representation over different shapes and textures: the linear relationship between horizontal and vertical derivatives. The code will be publicly released.

\end{abstract}

%%%%%%%%% BODY TEXT
\section{Introduction}
\label{sec:intro}

\begin{figure}[t]
	\begin{center}
		\includegraphics[width=1.0\linewidth]{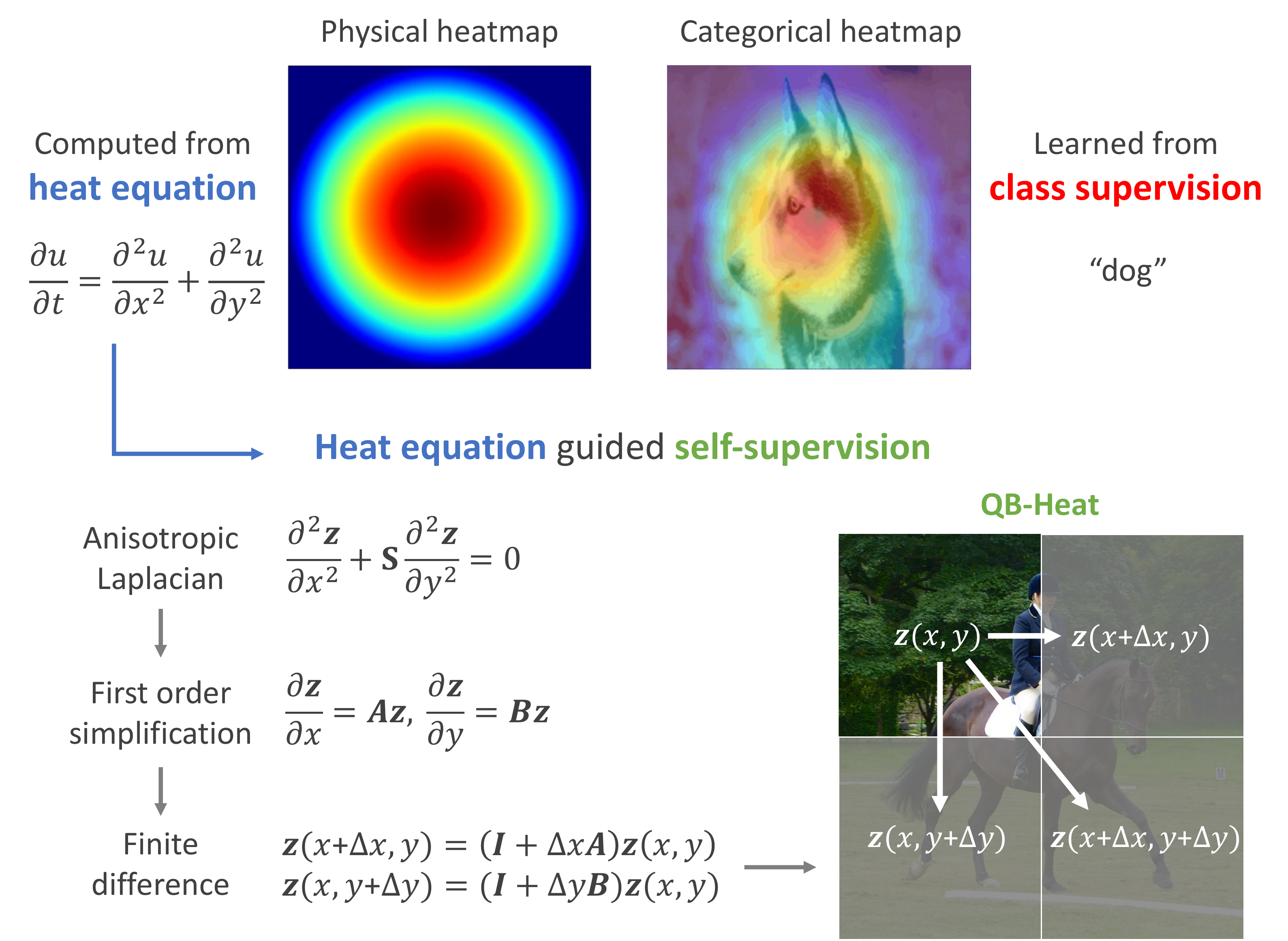}
	\end{center}
	\vspace{-4mm}
	\caption{\textbf{Overview of heat equation guided self-supervised learning.} Motivated by the connection between the physical heatmap computed from heat equation and the categorical heatmap (e.g. dog) learned from supervised learning, we leverage {\color{RoyalBlue}\textbf{heat equation}} to shift the learning from \textcolor{red}{\textbf{supervised}} to {\color{LimeGreen}\textbf{self-supervised}}. By simplifying heat equation into first order linear differential equations followed by finite difference approximation, we develop a simple masked image modeling method (named {\color{LimeGreen}\textbf{QB-Heat}}) that encodes a single unmasked quarter-block to predict other quarters via linear prediction. Best viewed in color.}
	\label{fig:heapmap}
	\vspace{-3mm}
\end{figure}

Recent work in class activation maps (CAM) \cite{zhou2015cnnlocalization} shows that convolutional neural networks (CNNs) followed by global average pooling is able to learn  \textit{categorical heatmap} (see \cref{fig:heapmap}) from image level supervision, which is similar to 
\textit{physical heat diffusion} governed by heat equation as:
\begin{align}
 \textbf{Heat Equation:} \;\;\;\;\;\;\; \frac{\partial u}{\partial t}=\frac{\partial^2 u}{\partial x^2}+\frac{\partial^2 u}{\partial y^2}, \nonumber
%\label{eq:heat-eq}
\end{align}
where the change of temperature $u$ over time $t$ is related to the change over 2D space $x$, $y$. This motivates us to use \textbf{\textit{ heat equation instead of class labels to guide representation learning}}, thus providing a new perspective of self-supervised learning.

To achieve this, we extend heat equation from \textit{measurable scalar} (i.e. temperature $u$) to \textit{latent vector} (i.e. feature vector $\bm{z}$ with $C$ channels). Then we add steady-state condition $\frac{\partial \bm{z}}{\partial t}=0$ to remove time dependence, and extend 2D isotropic Laplacian into linearly correlated as follows:
\begin{align}
 \textbf{Anisotropic Laplacian} \;\;\;\;\;\; \frac{\partial^2 \bm{z}}{\partial x^2}+\bm{S}\frac{\partial^2 \bm{z}}{\partial y^2}=0, \nonumber
%\label{eq:heat-eq-2}
\end{align}
where $\bm{z}$ is feature map with $C$ channels, i.e. $\bm{z}(x,y)\in\mathbb{R}^C$, and $\bm{S}$ is a $C$$\times$$C$ matrix. Here $\bm{S}$ plays two roles: (a) handling \textit{nonequivalent} change over horizontal and vertical directions, and (b) encoding  
\textit{invariant} relationship between the second order of derivatives along $x$ and $y$ axes in the latent representation space. Furthermore, we decouple this spatial invariance along $x$ and $y$ axes separately to simplify the anisotropic Laplacian into two first order linear differential equations as follows:  
\begin{align}
\textbf{First order linear:} \;\;\;\; \frac{\partial \bm{z}}{\partial x}=\bm{A}\bm{z}, \;\; \frac{\partial \bm{z}}{\partial y}=\bm{B}\bm{z}, \nonumber
%\label{eq:pde-linear}
\end{align}
where $\bm{A}$ and $\bm{B}$ are invertible matrices with size $C$$\times$$C$ and $\bm{S}$$=$$-\bm{A}^2(\bm{B}^2)^{-1}$. 
\begin{figure*}[t]
	\begin{center}
		\includegraphics[width=0.95\linewidth]{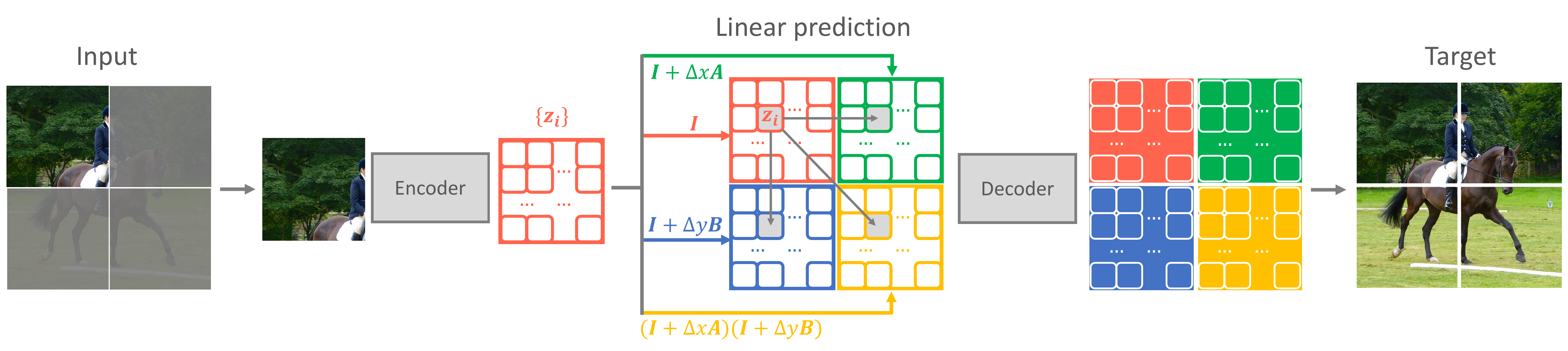}
	\end{center}
	\vspace{-4mm}
	\caption{\textbf{QB-Heat architecture.} An encoder is applied on a single \textit{unmasked} quarter-block to extract feature map $\{\bm{z}_i\}$, leaving other three quarters \textit{masked}. The prediction is performed between corresponding positions from unmasked to masked quarters through linear projection. A projection matrix is shared over all positions within each quarter, but three masked quarters have different projection matrices, i.e. $\bm{I}$$+$$\Delta x\bm{A}$, $\bm{I}$$+$$\Delta y\bm{B}$, and $(\bm{I}$$+$$\Delta x\bm{A})(\bm{I}$$+$$\Delta y\bm{B})$. Then a small decoder is used to reconstruct the original image in pixels. The self-prediction is based on the finite difference approximation, e.g. $\bm{z}(x$$+$$\Delta x, y)$$=$$(\bm{I}$$+$$\Delta x\bm{A})\bm{z}(x, y)$, where the spatial offsets are half width/height of the image, i.e. $\Delta x$$=$$W/2$, $\Delta y$$=$$H/2$. Best viewed in color.}
	\label{fig:overview}
	\vspace{-3mm}
\end{figure*}
This simplification not only has nice properties, like holding linear relationship for any order derivatives between $x$ and $y$ as $\frac{\partial^n \bm{z}}{\partial x^n}$$=$$\bm{A}^n(\bm{B}^m)^{-1}\frac{\partial^m \bm{z}}{\partial y^m}$, but also allows horizontal and vertical prediction based on its finite difference approximation as follows:
\begin{align}
\textbf{Finite} \;\;\;\;\;\;\;\; \bm{z}(x+\Delta x, y)-\bm{z}(x, y)&=\Delta x\bm{A}\bm{z}(x,y) \nonumber \\
\textbf{difference} \;\;\;\; \bm{z}(x, y+\Delta y)-\bm{z}(x, y)&=\Delta y\bm{B}\bm{z}(x,y).\nonumber
%\label{eq:pde-finite} 
\end{align}

This gives rise to a new masked image modeling method. Specifically, only a single quarter-block is \textit{unmasked} to encode $\bm{z}(x, y)$, which is used to predict other three \textit{masked} quarters $\bm{z}(x$$+$$\Delta x, y)$, $\bm{z}(x, y$$+$$\Delta y)$, $\bm{z}(x$$+$$\Delta x, y$$+$$\Delta y)$ via linear prediction (see \cref{fig:heapmap}, \ref{fig:overview}). The learning target includes encoder $\bm{z}$ and matrices $\bm{A}$, $\bm{B}$.  We name this \underline{Q}uarter-\underline{B}lock prediction guided by \underline{Heat} equation as QB-Heat.
Compared to popular MAE \cite{MaskedAutoencoders2021}, it has four differences:
\begin{itemize}
    \setlength\itemsep{0.4em}
    \item \textbf{more regular} masking (a single unmasked quarter).
    \item \textbf{simpler} linear prediction.
    \item enabling masked image modeling for \textbf{efficient CNN} based architectures without bells and whistles.
    \item modeling \textbf{spatial invariance} explicitly in representation space via learnable matrices $\bm{A}$ and $\bm{B}$.
\end{itemize}

We also present an evaluation protocol, \textit{decoder probing}, in which the frozen pre-trained encoder (without fine-tuning) is evaluated over two tasks (image classification and object detection) with different decoders. Decoder probing includes widely used linear probing, but extends from it by adding non-linear decoders. 
It directly evaluates encoders as they are, complementary to fine-tuning that evaluates pre-trained models indirectly as initial weights.

QB-Heat brings masked image modeling to CNN based architectures, even for pre-training light-weight networks.
Moreover, the pre-trained encoders are suitable for both image classification and object detection without fine-tuning. For instance, when pre-training a Mobile-Former \cite{MobileFormer-2022-cvpr} with 5.8M parameters and 285M FLOPs, QB-Heat is on par with MoCo-v2 \cite{chen2020mocov2} in linear probing on ImageNet, but outperforms by a clear margin (65.6\% vs. 52.9\%) in non-linear decoder probing that adds a transformer block before the linear classifier. When transferring to object detection with frozen backbone, QB-Heat outperforms MoCo-v2 and supervised pre-training on ImageNet by 7.9 and 4.5 AP respectively.
In addition, we found that fine-tuning QB-Heat pre-trained encoders on ImageNet-1K alone introduces consistent gain on object detection, thus providing strong encoders shared by classification and detection tasks. For example, 82.5\% top-1 accuracy on ImageNet and 45.5 AP on COCO detection (using 100 queries in DETR framework) are achieved by sharing a Mobile-Former with 25M parameters and 3.7G FLOPs (similar to ResNet-50 and ViT-S). 

The solid performance demonstrates that the simplified heat equation (from anisotropic Laplacian to first order linear differential equations) sheds light on the spatial invariance of visual representation: horizontal and vertical partial derivatives are linearly correlated. We hope this will encourage exploration of principles in visual representations.

%------------------------------------------------------------------------
\section{Related Work}
\label{sec:related-work}
\noindent \textbf{Contrastive methods}
\cite{BeckerHinton92, Hadsell2006dimreduction, Oord2018RepresentationLW, wu2018unsupervised, he2019moco, chen2020simsiam, caron2021emerging} achieve significant progress recently. They are most applied to Siamese architectures \cite{chen2020simple, he2019moco, chen2020mocov2, chen2021mocov3} to contrast image similarity and dissimilarity and rely on data augmentation. \cite{chen2020simsiam, grill2020bootstrap} remove dissimilarity between negative samples by handling collapse carefully. 
\cite{chen2020big, Li_2021_ICCV} show pre-trained models work well for semi-supervised learning and few-shot transfer.

\vspace{1mm}
\noindent \textbf{Information maximization} provides another direction to prevent collapse.  W-MSE \cite{ermolov2021whitening} avoids collapse by scattering batch samples to be uniformly distributed on a unit sphere. 
Barlow Twins \cite{zbontar2021barlow} decorrelates embedding vectors from two branches by forcing cross-correlation matrix to identity. VICReg \cite{bardes2022vicreg} borrows decorrelation mechanism from Barlow Twins, but explicitly adds variance-preservation for each variable of two embeddings. 

\vspace{1mm}
\noindent \textbf{Masked image modeling} (MIM) is inspired by the success of BERT \cite{devlin-etal-2019-bert} and ViT \cite{dosovitskiy2021vit} to learn representation by predicting masked region from unmasked counterpart. BEiT \cite{beit} and PeCo \cite{Dong2021Peco} predict on tokens, MaskFeat \cite{Wei_2022_CVPR_MaskFeat} predicts on HOG, and MAE \cite{MaskedAutoencoders2021} reconstructs original pixels. Recent works explore further improvement by combining MIM and contrastive learning \cite{zhou2021ibot, dong2022bootstrapped,huang2022cmae,tao-sim-2022,assran2022masked} or techniques suitable for ConvNets \cite{gao2022convmae,Li2022mscn,Fang2022CorruptedIM}. Different from these works that rely on random masking or ViT, our QB-Heat uses regular masking and simpler linear prediction to enable MIM for efficient CNNs without bells and whistles.

%------------------------------------------------------------------------
%\clearpage
\section{Heat Equation in Feature Space}
In this section, we discuss in details how to extend heat equation $\frac{\partial u}{\partial t}=\frac{\partial^2 u}{\partial x^2}+\frac{\partial^2 u}{\partial y^2}$ from a \textit{uni-dimensional and observed} variable (i.e. temperature $u$) into \textit{multi-dimensional and latent} feature space $\bm{z}$. 

\vspace{1mm}
\noindent \textbf{Motivation:} 
Motivated by class activation maps (CAM) \cite{zhou2015cnnlocalization} in which the categorical heatmap is similar to physical heat diffusion (see \cref{fig:heapmap}), we hypothesize that (a) the feature map around a visual object is smooth and governed by heat equation, and (b) the corresponding feature encoder can learn from heat equation alone without any labels. These hypotheses are hard to prove, but instead we show their potential in self-supervised learning. Next, we discuss how to extend heat equation into feature space.

\vspace{1mm}
\noindent \textbf{Extending heat equation into linear systems:}
The extension of heat equation is based on two design guild-lines: (a) the heat diffusions along multiple feature channels are correlated, and (b) the diffusions along horizontal and vertical directions are not equivalent. The former is straightforward as most of neural architectures output highly correlated features. The latter is due to the shape and texture \textit{anisotropy} in visual objects which determines the heat diffusions along features. This is different with original heat equation which is spatial \textit{isotropy} on a single channel.

Based on these two guild-lines, we firstly replace temperature $u$ in the original heat equation with feature vector $\bm{z}\in\mathbb{R}^C$ and use the steady-state condition $\frac{\partial \bm{z}}{\partial t}=0$ to remove time dependence, resulting in a Laplacian equation $\frac{\partial^2 \bm{z}}{\partial x^2}+\frac{\partial^2 \bm{z}}{\partial y^2}=0$. Then, we extend Laplacian from spatial isotropy to anisotropy by adding a coefficient matrix $\bm{S}$ with size $C\times C$ as $\frac{\partial^2 \bm{z}}{\partial x^2}+\bm{S}\frac{\partial^2 \bm{z}}{\partial y^2}=0$.  
To allow self-prediction along horizontal and vertical directions, we decouple $x$ and $y$ axes in Laplacian into two first-order linear differential equations as:
\begin{align}
\frac{\partial \bm{z}}{\partial x}=\bm{A}\bm{z}, \;\; \frac{\partial \bm{z}}{\partial y}=\bm{B}\bm{z},
\;\;\;\; \bm{S}=-\bm{A}^2(\bm{B}^2)^{-1},
\label{eq:pde-linear}
\end{align}
where $\bm{A}$ and $\bm{B}$ are two $C\times C$ matrices.
Note that $\bm{A}$ and $\bm{B}$ are commuting matrices $\bm{AB}=\bm{BA}$ if $\bm{z}(x,y)$ has continuous second partial derivatives based on the Clairaut's theorem ($\frac{\partial^2 \bm{z}}{\partial x\partial y}=\bm{BAz}=\bm{ABz}=\frac{\partial^2 \bm{z}}{\partial y\partial x}$). Here, we assume $\bm{A}$ and $\bm{B}$ are invertible matrices to achieve $\bm{S}$.

\vspace{2mm}
\noindent \textbf{Properties:}
The first-order simplification above is a special case of Laplacian that has nice properties as follows.

\vspace{1mm}
\noindent \textit{Property 1: linear relationship holds for any order derivatives between horizontal and vertical directions as:}
\begin{align}
&\frac{\partial^n \bm{z}}{\partial x^n}=\bm{A}^n\bm{z}, \;\;\;\; \frac{\partial^m \bm{z}}{\partial y^m}=\bm{B}^m\bm{z}, \nonumber  \\
&\frac{\partial^n \bm{z}}{\partial x^n}-\bm{A}^n(\bm{B}^m)^{-1}\frac{\partial^m \bm{z}}{\partial y^m}=0.
\label{eq:pde-norder}
\end{align}

\vspace{1mm}
\noindent \textit{Property 2 -- solution has exponential format as:}
\begin{align}
\bm{z}(x,y)=e^{\bm{A}x}e^{\bm{B}y}\bm{z}(0,0),
\label{eq:exp-matrix}
\end{align}
where the exponential matrix $e^{\bm{A}x}$ is defined by Taylor expansion $e^{\bm{A}x}$$=$$\sum_{n=0}^{\infty}(\bm{A}x)^n/n!$, and $\bm{z}(0,0)$ is the initial vector. Since $\bm{A}$ and $\bm{B}$ are commuting matrices (i.e. $\bm{AB}=\bm{BA}$), they share eigenvectors (denoted as $\bm{v}_i$) when $\bm{A}$ has distinct eigenvalues. Thus, \cref{eq:exp-matrix} can be written as:
\begin{align}
\bm{z}(x,y)=\sum_{i=1}^Cc_ie^{\lambda_ix+\pi_iy}\bm{v}_i,
\label{eq:solution}
\end{align}
where $\{\lambda_i\}$ and $\{\pi_i\}$ are eigenvalues for $\bm{A}$ and $\bm{B}$, respectively. The coefficient $c_i$ is determined by initial vector $\bm{z}(0,0)$ such that $\sum_ic_i\bm{v}_i=\bm{z}(0,0)$.

\vspace{1mm}
\noindent \textbf{From continuous to discrete:}
In practice, we approximate continuous coordinates ($x$, $y$)  by using discrete measure over $H \times W$ locations, converting \cref{eq:pde-linear} to difference over small segment ($\Delta x$ or $\Delta y$) as follows:
\begin{align}
\bm{z}(x+\Delta x, y)-\bm{z}(x, y)&=\Delta x\bm{A}\bm{z}(x,y) \nonumber \\
\bm{z}(x, y+\Delta y)-\bm{z}(x, y)&=\Delta y\bm{B}\bm{z}(x,y).
\label{eq:pde-finite}
\end{align}

\vspace{1mm}
\noindent \textbf{Collapse solution:}
Both continuous \cref{eq:pde-linear} and discrete \cref{eq:pde-finite} have a collapse solution, i.e. feature map has \textit{constant} value $\bm{z}(x,y)=c$, and $\bm{A}$ and $\bm{B}$ are \textit{zero} matrices. Inspired by LeCun's seminal paper \cite{JEPA2022} that discusses multiple ways to handle collapse, we propose a new masked image modeling guided by \cref{eq:pde-finite} to handle collapse. In particular, we mask out $(x$$+$$\Delta x, y)$ and $(x, y$$+$$\Delta y)$ and predict their features from \textit{unmasked} $\bm{z}(x, y)$ using linear projection in \cref{eq:pde-finite}. This new self-supervised pre-training based on linear differential equations is named QB-Heat, which will be discussed in details next.

\section{QB-Heat}
We now introduce \underline{Q}uarter-\underline{B}lock prediction guided by \underline{Heat} equation (QB-Heat), that performs self-prediction based on \cref{eq:pde-finite}. It not only prevents collapse but also enables masked image modeling for CNN based architectures.  

\begin{figure}[t]
	\begin{center}
		\includegraphics[width=1.0\linewidth]{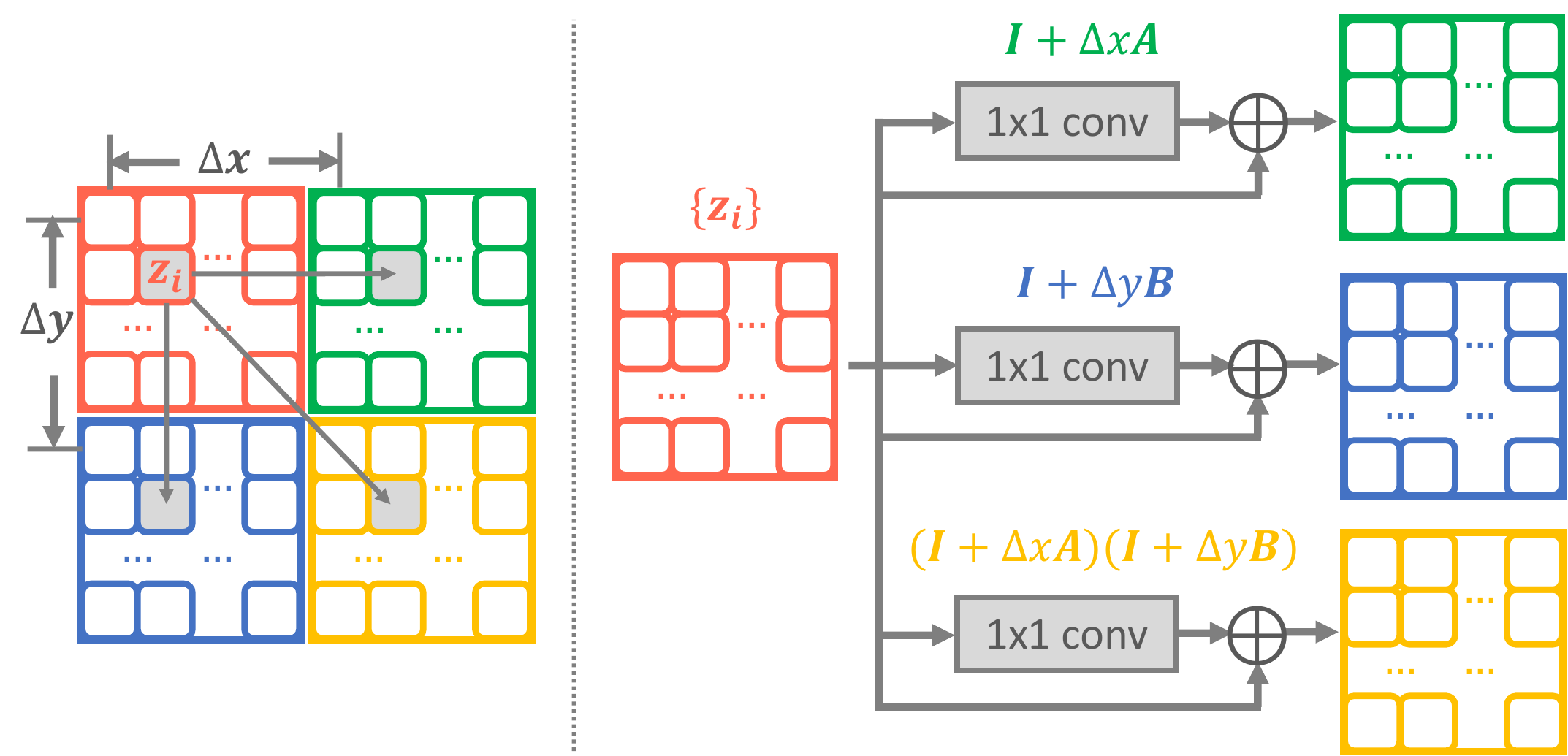}
	\end{center}
	\vspace{-4mm}
	\caption{\textbf{Implementation of translational linear prediction.} The feature map extracted from the unmasked top-left quarter of the input image is used to predict feature maps for other three masked quarters using 1$\times$1 convolution. Best viewed in color.}
	\label{fig:linear-pred}
	\vspace{-4mm}
\end{figure}
\subsection{Linear Prediction based on Quarter Masking}
 QB-Heat only uses a \textit{single unmasked} block to extrapolate over masked area via linear prediction. This resolves the conflict between random masking and CNN based encoder. The unmasked block has quarter size of the input image (see \cref{fig:overview}) and goes through encoder to extract features. Then, linear prediction is performed over three \textit{masked} quarter-blocks followed by a decoder to reconstruct the original image. The linear prediction is element-wise and can be implemented as 1$\times$1 convolution (see \cref{fig:linear-pred}). Each masked quarter-block has its own linear model, which is shared by all elements within the block. QB-Heat has two components to adjust: (a) the position of the unmasked quarter-block, and (b) the number of explicit linear models, which are discussed below.

\begin{figure}[t]
	\begin{center}
		\includegraphics[width=0.85\linewidth]{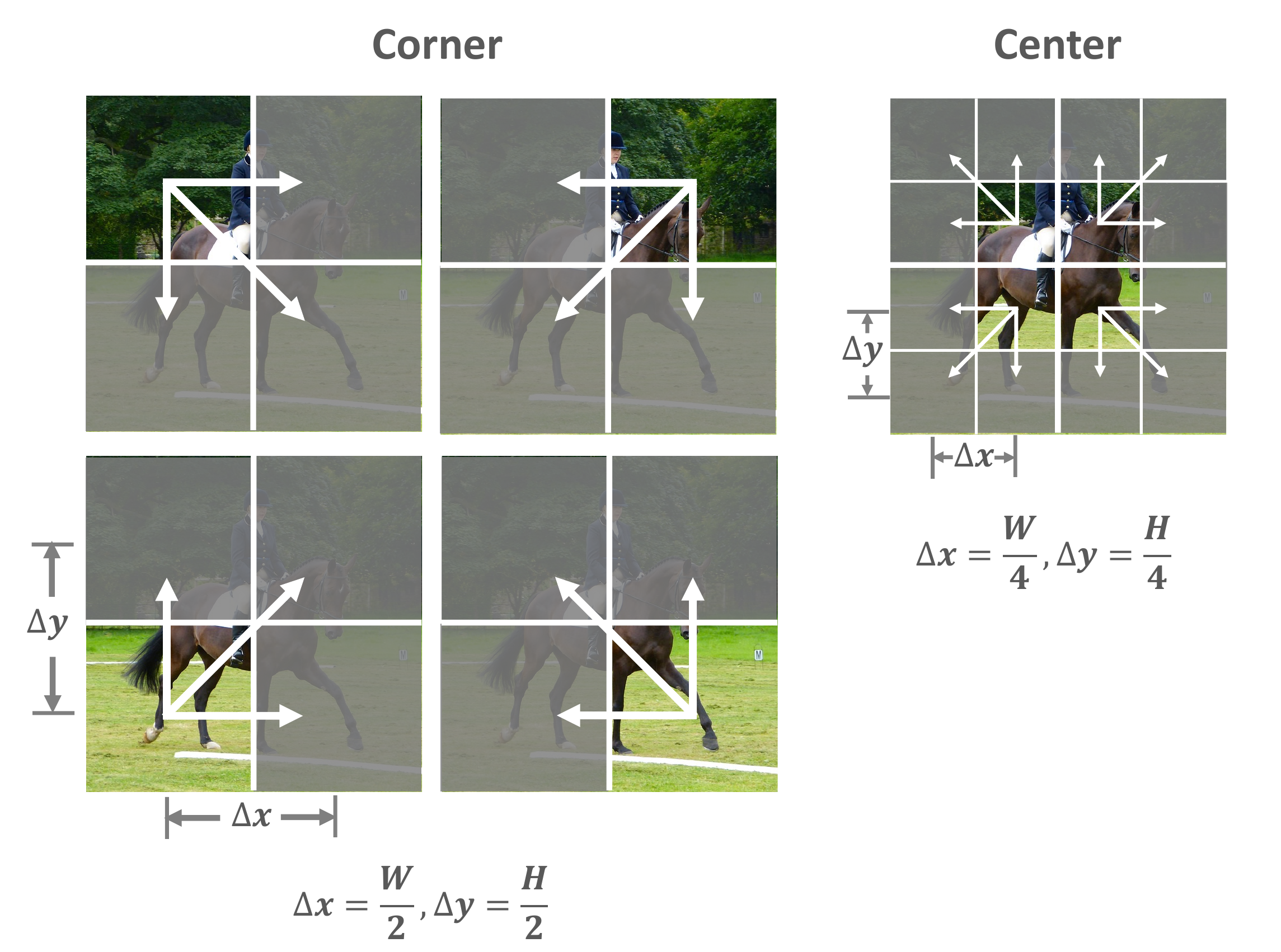}
	\end{center}
	\vspace{-4mm}
	\caption{\textbf{Position of the unmasked quarter-block}. The corner position corresponds to prediction over larger translation ($\Delta x$$=$$W/2$, $\Delta y$$=$$H/2$) than the center position ($\Delta x$$=$$W/4$, $\Delta y$$=$$H/4$).  }
	\label{fig:unmask-pos}
	\vspace{-3mm}
\end{figure}

\vspace{1mm} \noindent
\textbf{Position of unmasked quarter-block:}
The unmasked quarter-block are either at four corners or at the center (see \cref{fig:unmask-pos}), corresponding to prediction at different translation scales. Placing the unmasked quarter at \textit{corner} corresponds to a larger prediction offset $\Delta x$$=$$W/2$, $\Delta y$$=$$H/2$, while prediction from the \textit{center} quarter-block is at a finer scale $\Delta x$$=$$W/4$, $\Delta y$$=$$H/4$ after splitting it into four sub-blocks. Our experiments show that mixing corner and center positions in a batch provides the best performance.

\begin{figure}[t]
	\begin{center}
		\includegraphics[width=1.0\linewidth]{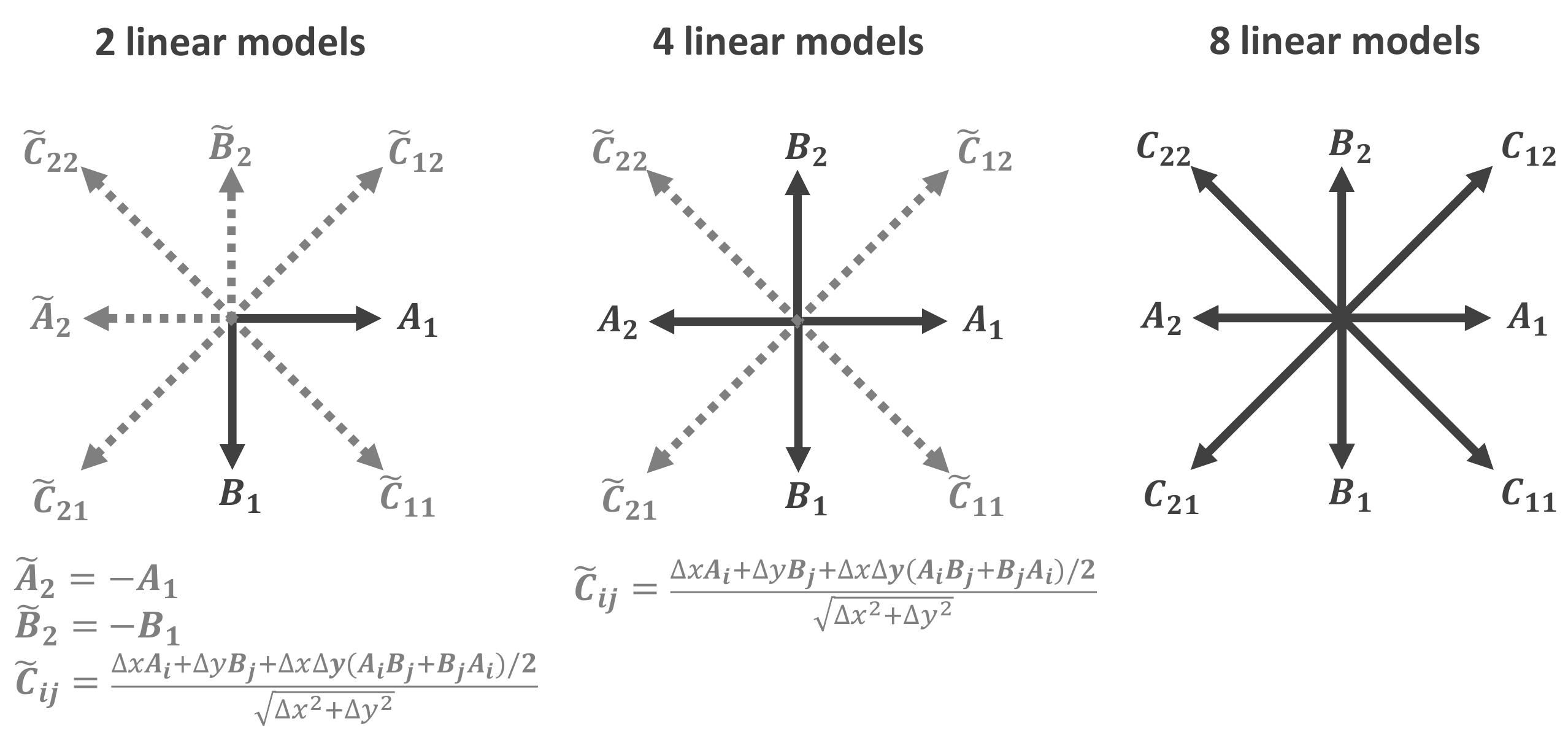}
	\end{center}
	\vspace{-4mm}
	\caption{\textbf{Number of explicit linear models.} A solid arrow indicates an explicit linear model, while a dash arrow indicates an implicit model derived from the explicit counterparts.}
	\label{fig:num-lin-models}
	\vspace{-3mm}
\end{figure}

\vspace{1mm} \noindent
\textbf{Number of explicit linear models:}
As shown in \cref{fig:unmask-pos}, prediction across blocks is performed along 8 directions in total for both corner and center positioning of the unmasked quarter. Two of them (right, down) are included in difference equations ($\bm{A}$, $\bm{B}$ in \cref{eq:pde-finite}). The other six can be either derived from $\bm{A}$ and $\bm{B}$ (see \cref{app:derivation-linear} for details) or modeled explicitly by adding linear models. \cref{fig:num-lin-models} shows three variants that have 2, 4 and 8 explicit linear models (solid arrow) respectively. The remaining directions (dash arrow) are derived from explicit models. Experiments show two explicit models work well, demonstrating $\bm{A}$ and $\bm{B}$ effectively encode the feature change.

\subsection{Architecture and Implementation}
QB-Heat follows masked autoencoder \cite{MaskedAutoencoders2021} architecture that includes masking, encoder, predictor and decoder. 

\vspace{1mm} \noindent
\textbf{Masking:}
QB-Heat has a single \textit{unmasked} block with quarter size of the input image, which is located at either corners or center (see \cref{fig:unmask-pos}). This is consistent with MAE in masking ratio (75\%), but is applicable for CNN based encoder.

\vspace{1mm} \noindent
\textbf{QB-Heat encoder:}
We use Mobile-Former \cite{MobileFormer-2022-cvpr} as encoder, which is a CNN based network (adding 6 global tokens in parallel to MobileNet \cite{sandler2018mobilenetv2}). To retain more spatial details, we increase the resolution for the last stage from $\frac{1}{32}$ to $\frac{1}{16}$. Three Mobile-Former variants (with 285M, 1.0G, 3.7G FLOPs) are used for evaluation. All of them has 12 blocks and 6 global tokens (see \cref{table:mf-enc-achs} in \cref{apx:arch-imgnet}).

\vspace{1mm} \noindent
\textbf{QB-Heat predictor:} 
The output features of the unmasked quarter-block are projected to 512 dimensions and followed by linear models (implemented as 1$\times$1 convolution in \cref{fig:linear-pred}) to predict for masked blocks. This predictor is only used in pre-training and removed during inference.

\vspace{1mm} \noindent
\textbf{QB-Heat decoder:}
We follow MAE \cite{MaskedAutoencoders2021} to apply a series of transformer blocks as decoder on both unmasked and masked quarter-blocks. In this paper, we use 6 transformer blocks with dimension 512 in decoder.

\subsection{Relation to MAE}
QB-Heat differentiates from MAE \cite{MaskedAutoencoders2021} by explicitly modeling feature derivatives using linear differential equations, enabling \textit{more regular} masking and  \textit{simpler} prediction to support \textit{more efficient} CNN based networks.

\vspace{1mm} \noindent
\textbf{More regular masking:}
Different with random unmasked patches in MAE, QB-Heat has a \textit{single unmasked quarter-block}, suitable for CNNs without bells and whistles. Compared to MAE with regular block-wise masking that achieves 63.9\% in linear probing and 82.8\% in fine-tuning on ImageNet-1K by using ViT-L with 307M parameters, QB-Heat achieves similar performance (65.1\% in linear probing, 82.5\% in fine-tuning) more efficiently by using Mobile-Former-3.7G with 35M parameters. 

\vspace{1mm} \noindent
\textbf{Simpler prediction:}
In QB-Heat, each masked patch is predicted from a \textit{single} unmasked patch with translation $\Delta x$ or $\Delta y$ (see \cref{fig:linear-pred}) %based on \cref{eq:pde-finite}.
rather than aggregating \textit{all} unmasked patches in MAE, thus resulting in much lower complexity.

\section{Evaluation: Decoder Probing}
\label{sec:eval}

In this section, we propose a new evaluation protocol for self-supervised pre-training to complement widely used linear probing and fine-tuning. Linear probing is sensitive to feature dimension and misses the opportunity of pursuing \textit{non-linear} features \cite{MaskedAutoencoders2021}, while fine-tuning \textit{indirectly} evaluates a pre-trained model as initial weights for downstream tasks. We need a new protocol that (a) can handle both linear and non-linear features, (b) performs direct evaluation without fine-tuning, (c) covers multiple visual tasks. It encourages exploration of pre-training a universal (or task-agnostic) encoder.

Decoder probing provides a solution. It involves multiple tasks such as image classification and object detection. For each task, only the decoder is learnable while the pre-trained encoder (backbone) is frozen. Each task has a set of decoders with different complexities to provide comprehensive evaluation. Below we list decoders used in this paper.

\vspace{1mm} \noindent
\textbf{Classification decoders:} We use two simple classification decoders: (a) \textit{linear decoder} (or linear probing) including global average pooling and a linear classifier, and (b) \textit{transformer decoder} that adds a single transformer block before global pooling (denoted as \texttt{tran-1}). The transformer block is introduced to encourage representative features that are not ready to separate categories linearly yet, but can achieve it by the assistance of a simple decoder.  

\vspace{1mm} \noindent
\textbf{Detection decoders:}
We use three detection decoders: two DETR \cite{nicolas2020detr} heads and one RetinaNet \cite{Lin_2017_ICCV_retinanet_focal} head. The two DETR heads use Mobile-Former \cite{MobileFormer-2022-cvpr} over three scales ($\frac{1}{32}$, $\frac{1}{16}$, $\frac{1}{8}$) with different depths. The shallower one (denoted as \texttt{MF-Dec-211}) has four blocks (two in $\frac{1}{32}$, one in $\frac{1}{16}$, one in $\frac{1}{8}$), while the deeper one (denoted as \texttt{MF-Dec-522}) has nine blocks (five in $\frac{1}{32}$, two in $\frac{1}{16}$, two in $\frac{1}{8}$). Please see \cref{table:od-mf-dec-achs} in \cref{apx:arch-coco} for details.
%------------------------------------------------------------------------
%%%%%%%%%%% table moco %%%%%%%%%%%%%
\begin{table}[t]
\parbox{.49\linewidth}{
	\begin{center}
	    \footnotesize
	    \setlength{\tabcolsep}{0.3mm}{
		\begin{tabular}{l|rr|c}
		    \specialrule{.1em}{.05em}{.05em} 
			 encoder & madds & param & \texttt{lin} \\
			\specialrule{.1em}{.05em}{.05em} 
			Mob-v3 \cite{Howard_2019_ICCV_mbnetv3}$^\dag$& 217M & 5.4M & 36.3 \\
			Eff-b0 \cite{tan-ICML19-efficientnet}$^\dag$ & 390M & 5.3M & 42.2 \\
			Eff-b1 \cite{tan-ICML19-efficientnet}$^\dag$& 700M & 7.8M & 50.7 \\
			MF-285M \cite{MobileFormer-2022-cvpr}$^\ddag$& 285M & 5.8M & \textbf{51.6} \\
			\specialrule{.1em}{.05em}{.05em} 
		\end{tabular}
		}
	\end{center}
	}
\hfill
\parbox{.48\linewidth}{
    \begin{center}
	    \footnotesize
	    \setlength{\tabcolsep}{0.3mm}{
		\begin{tabular}{l|rr|c}
		    \specialrule{.1em}{.05em}{.05em} 
			 encoder & madds & param & \texttt{lin} \\
			\specialrule{.1em}{.05em}{.05em} 
			
			Res-18 \cite{he2016deep}$^\dag$ & 1.8G & 11.7M & 52.5 \\
			Res-34 \cite{he2016deep}$^\dag$ & 3.6G & 21.8M & 57.4 \\
            MF-1.0G \cite{MobileFormer-2022-cvpr}$^\ddag$ & 1.0G& 13.5M & \textbf{60.4}\\ 
			\specialrule{.1em}{.05em}{.05em} 
			\multicolumn{4}{c}{}\\
		\end{tabular}
		}
	\end{center}
}
	\vspace{-4mm}
	\caption{\textbf{Linear probing results} of efficient networks pre-trained by MoCo-v2 \cite{chen2020mocov2}. ``MF" (e.g. MF-285M) refers to Mobile-Former. $^\dag$ and $^\ddag$ indicate implementation in \cite{fang2020seed} and this paper respectively.}
	\label{table:mocov2}
	\vspace{-4mm}
\end{table}

%%%%%%%%%%% table ablation %%%%%%%%%%%%%
\begin{table*}[t]
\begin{minipage}[b]{0.56\linewidth}
\parbox{.61\linewidth}{
%\centering
    \begin{center}
	    \footnotesize
		\setlength{\tabcolsep}{0.4mm}{
		\begin{tabular}{c|ccc}
		    \specialrule{.1em}{.05em}{.05em} 
			position (prediction offset) & \texttt{lin} & \texttt{tran-1} & \texttt{ft} \\
		
			\specialrule{.1em}{.05em}{.05em} 
			corner ($\Delta x$$=$$W/2, \Delta y$$=$$H/2$)  & 64.1 & 77.9 &  82.1  		 \\
			center ($\Delta x$$=$$W/4, \Delta y$$=$$H/4$) & 64.2 & 77.9 &  82.4  		 \\
			corner + center  & \cellcolor{lightgray}\textbf{65.1} & \cellcolor{lightgray}\textbf{78.6} & \cellcolor{lightgray} \textbf{82.5}   		 \\
			\specialrule{.1em}{.05em}{.05em} 
			\multicolumn{4}{c}{}\\[-0.5em]
			\multicolumn{4}{c}{(a) \textbf{Position of the unmasked quarter-block.}} \\
		\end{tabular}
		}
	\end{center}
	\label{table:ablation-masking}
	\vspace{-1.5em}
    
}
\hfill
\parbox{.37\linewidth}{
    \begin{center}
	    \footnotesize
	    \setlength{\tabcolsep}{0.45mm}{
		\begin{tabular}{c|ccc}
		    \specialrule{.1em}{.05em}{.05em} 
			\#models & \texttt{lin} & \texttt{tran-1} & \texttt{ft} \\
		
			\specialrule{.1em}{.05em}{.05em} 
			2  &  64.8 &  78.4 &  82.3  		 \\
			4  &  65.0 &  78.5 &  82.4		 \\
			8  & \cellcolor{lightgray}\textbf{65.1} & \cellcolor{lightgray}\textbf{78.6} & \cellcolor{lightgray}82.5  		 \\
			\specialrule{.1em}{.05em}{.05em} 
			\multicolumn{4}{c}{}\\[-0.5em]
			\multicolumn{4}{c}{(b) \textbf{Number of linear models.}} \\
		\end{tabular}
		}
	\end{center}
	\label{table:ablation-number-of-models}
	\vspace{-1.5em}
}
\caption{\textbf{QB-Heat ablation experiments} with Mobile-Former-3.7G on ImageNet-1K. We report top-1 accuracy (\%) of two decoder probings, i.e. linear (\texttt{lin}) and transformer (\texttt{tran-1}), and fine-tuning with transformer decoder  (\texttt{ft}). Two properties are observed: (a) multi-scale prediction (corner+center) is better than single scale, and (b) two explicit linear models ($\bm{A}$, $\bm{B}$ in \cref{eq:pde-finite}) are good enough.  Default settings are marked in \colorbox{lightgray}{gray}.}
\label{table:ablation}
\end{minipage}
	\quad
	\begin{minipage}[b]{0.41\linewidth}
	\begin{center}
		\includegraphics[width=1.0\linewidth]{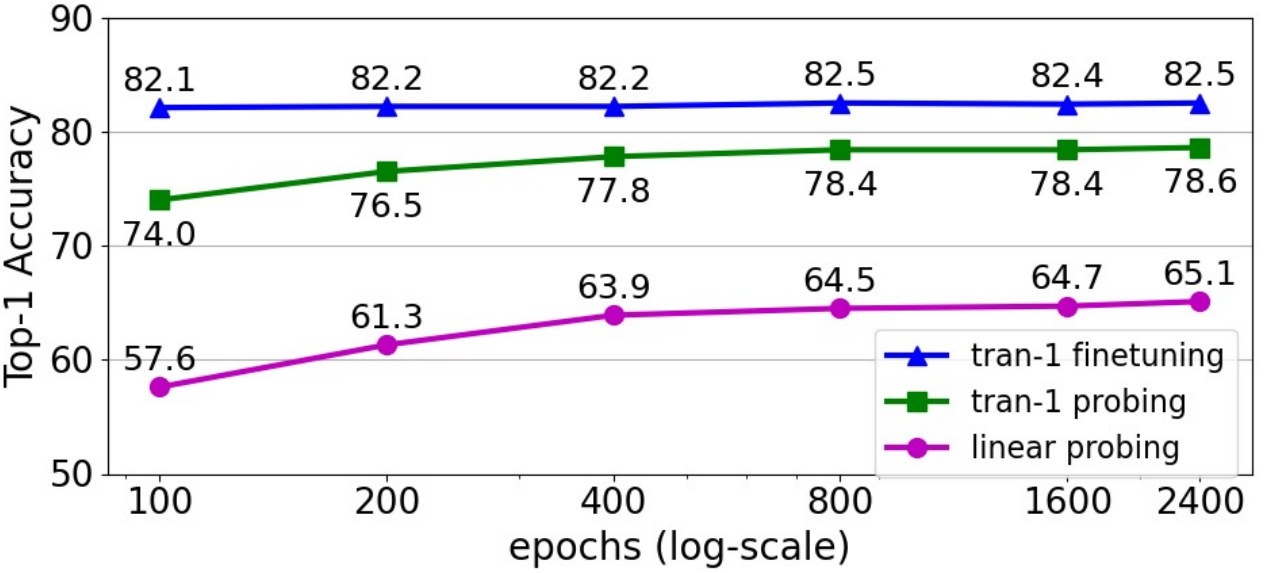}
	\end{center}
	\vspace{-4mm}
	\captionof{figure}{\textbf{Training schedules.} Longer training provides consistent improvement for linear and \texttt{tran-1} probing, while fine-tuning is not sensitive to training schedule. }
	\label{fig:linprob-epochs}
	\end{minipage}
\vspace{-2mm}
\end{table*}

\begin{figure}[t]
	\begin{center}
		\includegraphics[width=1.0\linewidth]{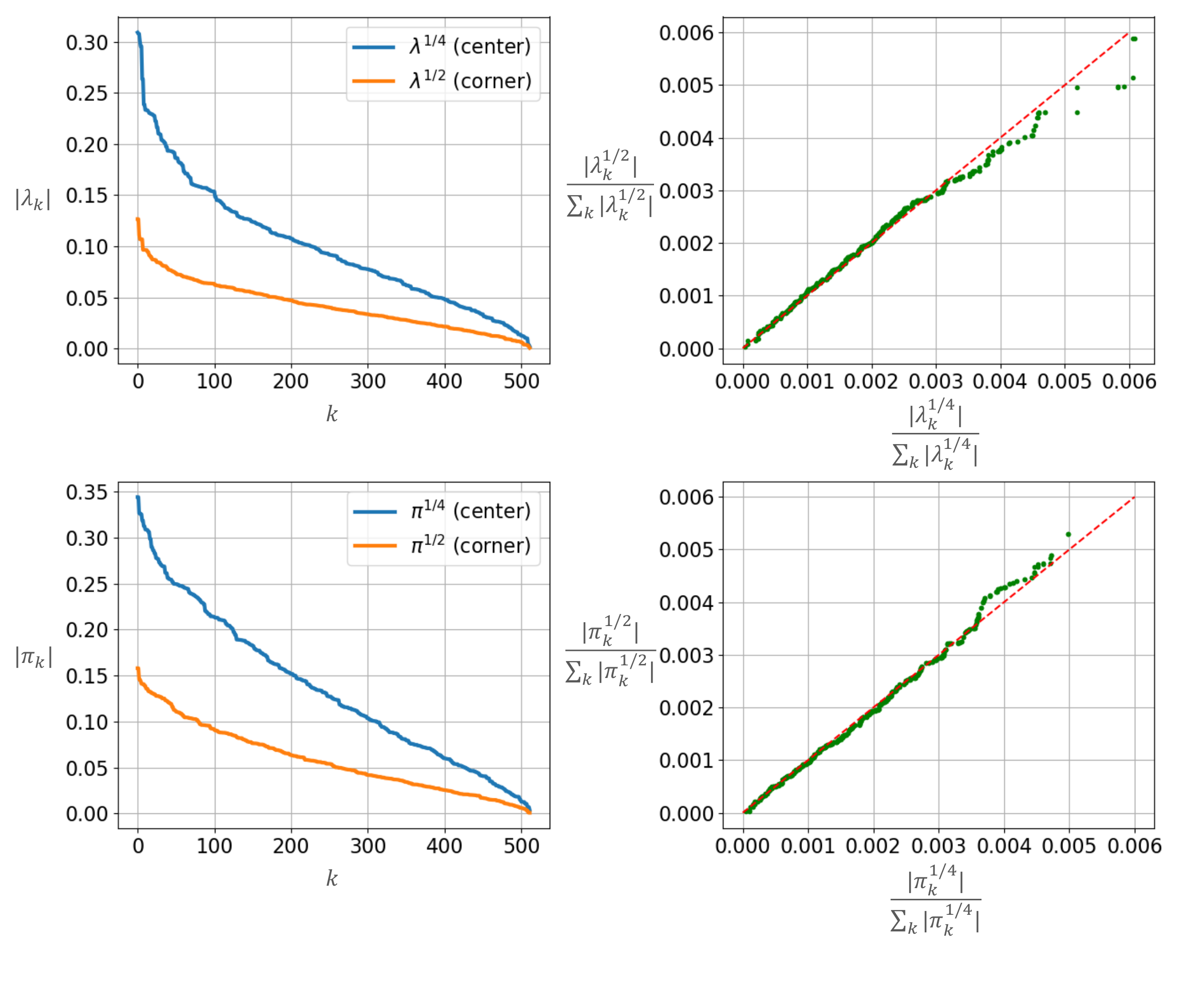}
	\end{center}
	\vspace{-9mm}
	\caption{\textbf{Spectrum of matrices $\bm{A}$ and $\bm{B}$} learned from QB-Heat pre-training on ImageNet-1K. Two sets of $\bm{A}$ and $\bm{B}$ are jointly learned at two scales in one batch. Half batch uses the center position of the unmasked quarter to predict over translation $\Delta x$$=$$W/4$, $\Delta y$$=$$H/4$, while the other half uses the corner position of the unmasked quarter to predict over translation $\Delta x$$=$$W/2$, $\Delta y$$=$$H/2$. We use $\bm{A}^{\frac{1}{4}}$,$\bm{B}^{\frac{1}{4}}$ and $\bm{A}^{\frac{1}{2}}$,$\bm{B}^{\frac{1}{2}}$ to denote matrices at these two scales respectively. Similar spectrum distribution is observed between $\bm{A}^{\frac{1}{4}}$ and $\bm{A}^{\frac{1}{2}}$ (and between $\bm{B}^{\frac{1}{4}}$ and $\bm{B}^{\frac{1}{2}}$). \textbf{Left column:} distribution of magnitude of eigenvalues ($\lambda_k$ and $\pi_k$ denotes the eigenvalues for $\bm{A}$ and $\bm{B}$ respectively). \textbf{Right column:} normalized eigenvalues across scales are aligned along the diagonal line. Best viewed in color.}
	\label{fig:spec}
\end{figure}

\section{Experiments}
We evaluate QB-Heat on both ImageNet-1K \cite{deng2009imagenet} and COCO 2017 \cite{lin2014microsoft}. CNN based Mobile-Former \cite{MobileFormer-2022-cvpr} is used as encoder as it outperforms other efficient CNNs in both supervised and self-supervised (see \cref{table:mocov2}) learning. Three variants with 285M, 1.0G and 3.7G FLOPs are used (see \cref{table:mf-enc-achs} in \cref{apx:arch-imgnet} for network details).

\vspace{1mm} \noindent
\textbf{ImageNet-1K \cite{deng2009imagenet}:} 
QB-Heat pre-training is performed on ImageNet-1K training set. Then, pre-trained encoders are \textit{frozen} and evaluated by two decoder probing (see \cref{sec:eval}): (a) linear probing, (b) \texttt{tran-1} probing that includes a single transformer block followed by a linear classifier. The fine-tuning performance of \texttt{tran-1} is also provided. Top-1 validation accuracy of a single 224×224 crop is reported.

\vspace{1mm} \noindent
\textbf{COCO 2017 \cite{lin2014microsoft}:} 
We also evaluate QB-Heat pre-training on COCO object detection that contains 118K training and 5K validation images. The \textit{frozen} encoders are evaluated using two decoders in DETR \cite{nicolas2020detr} framework. The training setup, fine-tuning performance and evaluation in RetinaNet \cite{Lin_2017_ICCV_retinanet_focal} are provided in \cref{apx:arch-coco,apx:more-exp}.

\subsection{Main Properties on ImageNet}

We ablate QB-Heat using the default setting in \cref{table:ablation} (see caption), and observe three properties listed below.

\vspace{1mm} \noindent
\textbf{Multi-scale prediction is better than single scale:}
\cref{table:ablation}-(a) studies the influence of the position of unmasked quarter-block. Placing the unmasked quarter at center or corner (see \cref{fig:unmask-pos}) corresponds to different scales of prediction offset in \cref{eq:pde-finite}, i.e. $\Delta x$$=$$W/2$, $\Delta y$$=$$H/2$ for corner position and $\Delta x$$=$$W/4$, $\Delta y$$=$$H/4$ for center position. Similar performance is achieved at either individual scale (center or corner position), while combining them in a batch (half for center and half for corner) achieves additional gain, indicating the advantage of multi-scale prediction.

\vspace{1mm} \noindent
\textbf{Two linear models ($\bm{A}$, $\bm{B}$) are good enough to predict over 8 directions:}
We compare different number of linear models in \cref{table:ablation}-(b). Using 8 explicit linear models along 8 directions (see \cref{fig:num-lin-models}) has similar performance to using 2 or 4 explicit models while approximating the rest of directions.

\vspace{1mm} \noindent
\textbf{Long training schedule helps more on decoder probing than fine-tuning:}
\cref{fig:linprob-epochs} shows the influence of the length of training schedule. The accuracies of two decoder probings (linear and \texttt{tran-1}) improve steadily as training lasts longer, while fine-tuning with \texttt{tran-1} achieves decent performance even on pre-training for 100 epochs. This is different from MAE \cite{MaskedAutoencoders2021}, in which fine-tuning relies on longer training to improve. Similar trend is observed in other two Mobile-Former variants (see \cref{fig:linprob-epochs-all} in \cref{apx:more-exp}).

\subsection{Interesting Observations in Matrices $A$, $B$}
\begin{table}[t!]
	\begin{center}
	    \footnotesize
	    \setlength{\tabcolsep}{5.5mm}{
		\begin{tabular}{r|cc }
		    \specialrule{.1em}{.05em}{.05em} 
			epoch & $E(\bm{A^{\frac{1}{4}}})/E(\bm{B^{\frac{1}{4}}})$ & $E(\bm{A^{\frac{1}{2}}})/E(\bm{B^{\frac{1}{2}}})$ \\
			\hline
			200 & 0.9388 & 0.9418 \\
            400 & 0.8575 & 0.8559 \\
            800 & 0.8104 & 0.7997 \\
            1600 & 0.7336 & 0.7217 \\
            2400 & 0.6859 & 0.6813 \\
			\specialrule{.1em}{.05em}{.05em} 
		\end{tabular}
		}
	\end{center}
	\vspace{-4mm}
	\caption{\textbf{Spectrum energy ratio} between horizontal and vertical matrices $\bm{A}$ and $\bm{B}$. Predictions at different scales ($\Delta x$$=$$W/4$, $\Delta y$$=$$H/4$ vs. $\Delta x$$=$$W/2$, $\Delta y$$=$$H/2$) have similar energy ratio, which becomes smaller as training schedule gets longer.}
	\label{table:energy-ratio}
\end{table}

Empirically, we observe interesting patterns in matrices $\bm{A}$ and $\bm{B}$ \textit{learned} from QB-Heat pre-training. $\bm{A}$ and $\bm{B}$ are coefficient matrices of linear differential equations \cref{eq:pde-linear} (our simplification  of heat equation). The experiment is set up as follows. We perform QB-Heat pre-training on ImageNet-1K by mixing two-scale prediction in a batch. Specifically, half batch uses the center position of the unmasked quarter to predict over translation $\Delta x$$=$$W/4$, $\Delta y$$=$$H/4$, while the other half uses the corner position of the unmasked quarter to predict over translation $\Delta x$$=$$W/2$, $\Delta y$$=$$H/2$. Each scale learns its own $\bm{A}$ and $\bm{B}$ (denoted as $\bm{A}^{\frac{1}{4}}$,$\bm{B}^{\frac{1}{4}}$ and $\bm{A}^{\frac{1}{2}}$,$\bm{B}^{\frac{1}{2}}$ respectively). All of them have dimension 512$\times$512. 

Three interesting patterns are observed in these learned matrices. Firstly, they have full rank with complex eigenvalues. Secondly, as shown in \cref{fig:spec}, $\bm{A}^{\frac{1}{4}}$ and $\bm{A}^{\frac{1}{2}}$ have similar spectrum distribution (magnitude of eigenvalues). Similarly, $\bm{B}^{\frac{1}{4}}$ and $\bm{B}^{\frac{1}{2}}$ have similar spectrum distribution. The right column of \cref{fig:spec} plots the sorted and normalized magnitude of eigenvalues (divided by the sum) between $\bm{A}^{\frac{1}{4}}$ and $\bm{A}^{\frac{1}{2}}$. They are well aligned along the diagonal red line. Thirdly, although $\bm{A}$ and $\bm{B}$ have different spectrum energy, their ratio is approximately scale invariant: 
\begin{align}
\frac{E(\bm{A^{\frac{1}{2}}})}{E(\bm{B^{\frac{1}{2}}})}\approx \frac{E(\bm{A^{\frac{1}{4}}})}{E(\bm{B^{\frac{1}{4}}})},
\label{eq:energy-ratio}
\end{align}
where the spectrum energy is computed as the sum of magnitude of eigenvalues as: 
\begin{align}
E(\bm{A})=\sum_{k=1}^n|\lambda_k|, \;\;\;\; E(\bm{B})=\sum_{k=1}^n|\pi_k|,
\label{eq:energy-def}
\end{align}
where $\lambda_k$ and $\pi_k$ are eigenvalues for $\bm{A}$ and $\bm{B}$ respectively. \cref{table:energy-ratio} shows that the energy ratio are approximately scale-invariant over different training schedules from 200 to 2400 epochs. The ratio reduces as training gets longer.

\subsection{Multi-task Decoder Probing}
\label{sec:multitask-decoder-probing-results}

%%%%%%%%%%% table tran-1 probing %%%%%%%%%%%%%
\begin{table*}[htb!]
\begin{minipage}[b]{0.28\linewidth}
	\begin{center}
	    \footnotesize
	    \setlength{\tabcolsep}{0.5mm}{
		\begin{tabular}{l|ccc}
		    \specialrule{.1em}{.05em}{.05em} 
			method & MF-285M & MF-1.0G & MF-3.7G \\
			\specialrule{.1em}{.05em}{.05em} 
			supervised & 75.7 & 79.4 & 80.8  		 \\
			MoCo-v2 & 74.3 & 79.2 & 80.0 \\
			\textbf{QB-Heat} & \textbf{75.8} & \textbf{80.5} & \textbf{82.5} \\
			\specialrule{.1em}{.05em}{.05em} 
		\end{tabular}
		}
	\end{center}
	\vspace{-3mm}
	\caption{\textbf{Fine-tuning results}, evaluated on ImageNet-1K. QB-Heat consistently outperforms baselines over three models. The gain increases as the model gets bigger. \texttt{tran-1} decoder is used for all methods.}
	\label{table:qbheat-vs-sup}
\end{minipage}
\quad
\begin{minipage}[b]{0.69\linewidth}
	\begin{center}
		\includegraphics[width=1.0\linewidth]{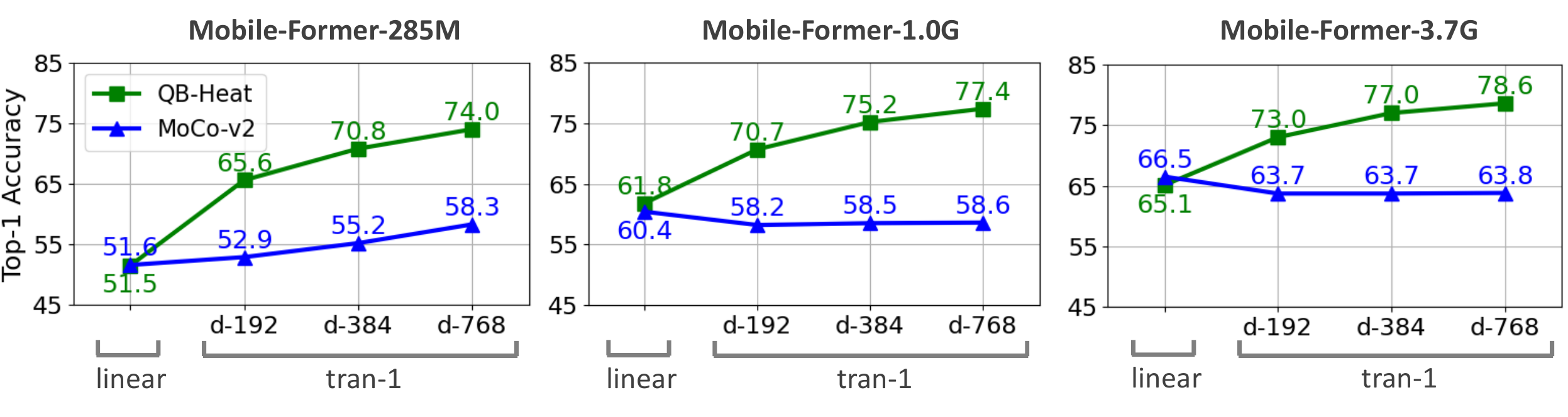}
	\end{center}
	\vspace{-4mm}
	\captionof{figure}{\textbf{Decoder probing on ImageNet-1K.} QB-Heat is on par with MoCo-v2 on linear probing, but outperforms on \texttt{tran-1} probing. The gain increases as the decoder gets wider. }
	\label{fig:in1k-decoder-probing}
\end{minipage}
	\vspace{-2mm}
\end{table*}

Here we report decoding probing results on both image classification and object detection. Each task includes multiple decoders. Note that the pre-trained encoders are \textit{\textbf{frozen}} even when transferring to COCO object detection.

\vspace{1mm} \noindent
\textbf{ImageNet classification:}
\cref{fig:in1k-decoder-probing} compares QB-Heat with MoCo-v2 \cite{chen2020mocov2} on linear and \texttt{tran-1} probing (see \cref{sec:eval}). When evaluating on \texttt{tran-1} probing, three widths (192, 384, 768) are used in the added transformer block. QB-Heat is on par with MoCo-v2 on linear probing, but is significantly better on \texttt{tran-1} probing. For instance, when using 192 channels in \texttt{tran-1} decoder to evaluate pre-trained Mobile-Former-285M, QB-Heat outperforms MoCo-v2 by 12.7\% (65.6\% vs. 52.9\%). This demonstrates that QB-Heat learns stronger non-linear spatial features. 

QB-Heat not only works well for decoder probing, but also provides a good initial for fine-tuning. As shown in \cref{table:qbheat-vs-sup}, its fine-tuning performance consistently outperforms the supervised counterpart over three models. The gain is larger for bigger models. Interestingly, fine-tuning on ImageNet-1K alone (freezing on COCO) boosts detection performance, providing strong task-agnostic encoders.

%%%%%%%%%%%%%%%%%%%%%%%%%%%%%%%%%%%%%%%%%
\begin{table*}[t]
    \begin{minipage}[b]{0.64\linewidth}
	\begin{center}
	    \smallfootnote
	    \setlength{\tabcolsep}{0.7mm}{
		\begin{tabular}{ccc|ccccc|lcc|ccc}
		    \specialrule{.1em}{.05em}{.05em} 
			& \textbf{head} & &
			\multicolumn{5}{c|}{\textbf{backbone}}&
			\multirow{3}{*}{AP} & 
			\multirow{3}{*}{AP\textsubscript{50}} & \multirow{3}{*}{AP\textsubscript{75}} & \multirow{3}{*}{AP\textsubscript{S}} & \multirow{3}{*}{AP\textsubscript{M}} & \multirow{3}{*}{AP\textsubscript{L}} \\
	         model & madds & param & model & madds & param & pre-train & IN-ft & &  & & &   \\
	         & (G) & (M) & & (G) & (M) & & & & & & & \\
	         
			\specialrule{.1em}{.05em}{.05em} 
			\multirow{15}{*}{\makecell{\texttt{MF}\\ \texttt{Dec} \\ \texttt{522}}} & \multirow{5}{*}{34.6} & \multirow{5}{*}{19.4} & \multirow{5}{*}{\makecell{MF\\3.7G}} & \multirow{5}{*}{77.5} & \multirow{5}{*}{25.0} & sup & -- & 40.5 & 58.5 & 43.3 & 21.1 & 43.4 & 56.8\\
			& & & & & & moco2 & \xmark     & 25.5\textsubscript{\textcolor{red}{(-15.0)}} & 40.4 & 26.7 & 12.3 & 27.2 & 37.0 \\
			& & & & & & moco2 & \checkmark          & 31.7\textsubscript{\textcolor{red}{(-8.8)}} & 48.3 & 33.5 & 16.1 & 33.4 & 45.5 \\
			& & & & & & \textbf{QB-Heat} & \xmark & 43.5 \textsubscript{\color{ForestGreen}{(+3.0)}}& 61.3 & 47.2 & 23.2 & 47.1 & 60.6 \\
			& & & & & & \textbf{QB-Heat} & \checkmark & \textbf{45.5} \textsubscript{\color{ForestGreen}{(+5.0)}}& \textbf{64.0} & \textbf{49.3} & \textbf{25.2} & \textbf{49.1} & \textbf{63.5} \\
			\cline{2-14}
			& \multirow{5}{*}{32.3} & \multirow{5}{*}{18.6} & \multirow{5}{*}{\makecell{MF\\1.0G}} 
            & \multirow{5}{*}{20.4} & \multirow{5}{*}{11.7} & sup & --& 38.3 & 56.0 & 40.8 & 19.0 & 40.9 & 54.3 \\
			& & & & & & moco2 & \xmark & 30.3\textsubscript{\textcolor{red}{(-8.0)}} & 46.0 & 32.3 & 15.1 & 32.1 & 42.5 \\
			& & & & & & moco2 & \checkmark & 39.0\textsubscript{\color{ForestGreen}{(+0.7)}} & 56.8 & 41.8 & 19.2 & 41.8 & 55.3 \\
			& & & & & & \textbf{QB-Heat} & \xmark & 42.6\textsubscript{\color{ForestGreen}{(+4.3)}} & 60.4 & 46.2 & 22.7 & 46.3 & 59.9 \\
			& & & & & & \textbf{QB-Heat} & \checkmark & \textbf{44.0}\textsubscript{\color{ForestGreen}{(+5.7)}} & \textbf{62.5}& \textbf{47.2} & \textbf{23.5} &\textbf{47.6} & \textbf{61.1} \\
			\cline{2-14}
			& \multirow{5}{*}{31.1} & \multirow{5}{*}{18.2} & \multirow{5}{*}{\makecell{MF\\285M}} 
            & \multirow{5}{*}{5.6} & \multirow{5}{*}{4.9} & sup & --& 35.2 & 52.1 & 37.6 & 16.9 & 37.2 & 51.7 \\
			& & & & & & moco2 & \xmark & 31.8\textsubscript{\textcolor{red}{(-3.4)}} & 47.8 & 34.1 & 14.9 & 33.3 & 45.6 \\
			& & & & & & moco2 & \checkmark & 39.9\textsubscript{\color{ForestGreen}{(+4.7)}} & 57.9 & 42.7 & 19.0 & 43.1 & 57.1 \\
			& & & & & & \textbf{QB-Heat} & \xmark & 39.7\textsubscript{\color{ForestGreen}{(+4.5)}} & 57.6 & 42.7 & 20.4 & 42.6 & 56.4 \\
			& & & & & & \textbf{QB-Heat} & \checkmark & \textbf{41.6}\textsubscript{\color{ForestGreen}{(+6.4)}} & \textbf{59.2} & \textbf{45.0} & \textbf{21.4} & \textbf{45.2} & \textbf{58.4} \\
			\specialrule{.1em}{.05em}{.05em}
			\multirow{15}{*}{\makecell{\texttt{MF} \\ \texttt{Dec} \\ \texttt{211}}} & \multirow{5}{*}{15.7} & \multirow{5}{*}{9.2} & \multirow{5}{*}{\makecell{MF\\3.7G}} & \multirow{5}{*}{77.5} & \multirow{5}{*}{25.0} & sup & -- & 34.1 & 51.3 & 36.1 & 15.5 & 36.8 & 50.0 \\
			& & & & & & moco2 & \xmark & 12.2\textsubscript{\textcolor{red}{(-21.9)}} & 24.1 & 10.7 & 5.3 & 13.0 & 19.3 \\
			& & & & & & moco2 & \checkmark & 19.1\textsubscript{\textcolor{red}{(-15.0)}} & 33.1 & 18.4 & 8.6&19.6 & 29.3\\
			& & & & & & \textbf{QB-Heat} & \xmark & 36.7\textsubscript{\color{ForestGreen}{(+2.6)}} & 53.8 & 39.5 & 17.2 & 39.6 & 53.5 \\
			& & & & & & \textbf{QB-Heat} & \checkmark & \textbf{41.0}\textsubscript{\color{ForestGreen}{(+6.9)}} & \textbf{59.3} & \textbf{44.2} & \textbf{20.9} & \textbf{44.5} & \textbf{58.2} \\
			\cline{2-14}
			& \multirow{5}{*}{13.4} & \multirow{5}{*}{8.4} & \multirow{5}{*}{\makecell{MF\\1.0G}} 
            & \multirow{5}{*}{20.4} & \multirow{5}{*}{11.7} & sup & -- & 31.2 & 47.8 & 32.8 & 13.7 & 32.9 & 46.9 \\
			& & & & & & moco2 & \xmark & 16.9\textsubscript{\textcolor{red}{(-14.3)}} & 29.7 & 16.4 & 7.7 & 17.6 & 25.8 \\
			& & & & & & moco2 & \checkmark & 30.6\textsubscript{\textcolor{red}{(-0.6)}}&46.7 & 32.1& 14.4& 32.0& 45.2\\
			& & & & & & \textbf{QB-Heat} & \xmark & 35.7\textsubscript{\color{ForestGreen}{(+4.5)}} & 52.5 & 38.5 & 16.9 & 38.6 & 51.5 \\
			& & & & & & \textbf{QB-Heat} & \checkmark & \textbf{39.3}\textsubscript{\color{ForestGreen}{(+8.1)}} & \textbf{56.8} & \textbf{42.0} & \textbf{18.9} & \textbf{43.1} & \textbf{56.3} \\
			\cline{2-14}
			& \multirow{5}{*}{12.2} & \multirow{5}{*}{8.0} & \multirow{5}{*}{\makecell{MF\\285M}} 
            & \multirow{5}{*}{5.6} & \multirow{5}{*}{4.9} & sup & -- & 27.8 & 43.4 & 28.9 & 11.3 & 29.1 & 41.6 \\
			& & & & & & moco2 & \xmark & 22.1\textsubscript{\textcolor{red}{(-5.7)}} & 35.7 & 22.8 & 9.6 & 22.4 & 34.4\\
			& & & & & & moco2 & \checkmark & 32.7\textsubscript{\color{ForestGreen}{(+4.9)}} & 49.0 & 34.6 & 14.5 & 35.1 & 48.8\\
			& & & & & & \textbf{QB-Heat} & \xmark & 33.0\textsubscript{\color{ForestGreen}{(+5.2)}} & 49.3 & 35.1 & 15.6 & 35.2 & 48.5 \\
			& & & & & & \textbf{QB-Heat} & \checkmark & \textbf{35.8}\textsubscript{\color{ForestGreen}{(+8.0)}} & \textbf{52.8} & \textbf{38.3} & \textbf{16.5} & \textbf{38.4} & \textbf{51.5} \\
			
			\specialrule{.1em}{.05em}{.05em} 
		\end{tabular}
		}
	\end{center}
	\vspace{-3mm}
	\caption{\textbf{COCO object detection results} on \texttt{val2017} for \textbf{\textit{frozen}} backbone pre-trained on ImageNet-1K. Evaluation is conducted over three backbones and two heads that use Mobile-Former \cite{MobileFormer-2022-cvpr} end-to-end in DETR \cite{nicolas2020detr} framework. Our QB-Heat significantly outperforms MoCo-v2 and supervised baselines. Fine-tuning on ImageNet-1K provides consistent improvement. Initial ``MF" (e.g. \texttt{MF-Dec-522}) refers to Mobile-Former. ``IN-ft" indicates fine-tuning on ImageNet-1K. MAdds is based on the image size 800$\times$1333.}
	\label{table:coco-det-results}
	\end{minipage}
	\quad
	\begin{minipage}[b]{0.33\linewidth}
	\begin{center}
		\includegraphics[width=1.0\linewidth]{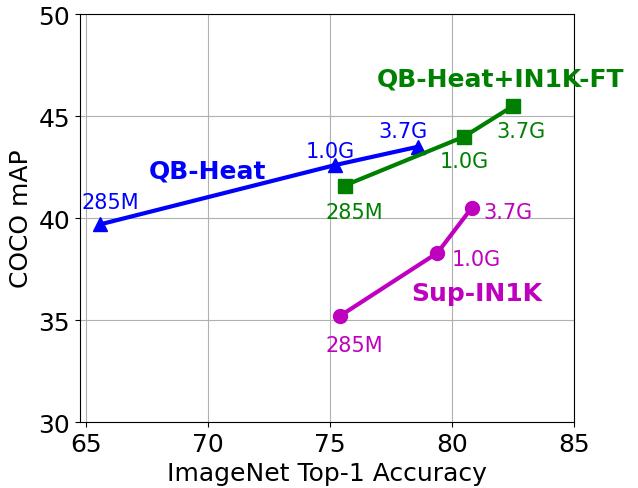}
	\end{center}
	\vspace{-4mm}
	\captionof{figure}{\textbf{Task-agnostic encoder,} evaluated on both ImageNet classification and COCO object detection. {\color{Mulberry}Sup-IN1K} indicates supervised pre-training on ImageNet-1K. \textcolor{blue}{QB-Heat} indicates QB-Heat pre-training while {\color{PineGreen}QB-Heat+IN1K-FT} indicates QB-Heat pre-training followed by fine-tuning on ImageNet-1K. For each pre-training, the three dots correspond to three Mobile-Former backbone variants. When evaluating on image classification, a \texttt{tran-1} decoder is added on top and learnt from class supervision. When evaluating on object detection, the nine layer head \texttt{MF-Dec-522} is used and the backbone is \textbf{\textit{frozen}}. Thus, the backbone is shared by classification and detection tasks. \textcolor{blue}{QB-Heat} is far behind {\color{Mulberry}Sup-IN1K} on image classification, but overtakes on object detection.  {\color{PineGreen}QB-Heat+IN1K-FT} boosts detection performance by fine-tuning on ImageNet-1K, providing strong task-agnostic encoders.}
	\label{fig:agnostic-enc}
    \end{minipage}
	\vspace{-2mm}
\end{table*}

\vspace{1mm} \noindent
\textbf{COCO object detection:}
\cref{table:coco-det-results} compares QB-Heat with MoCo-V2 and ImageNet supervised pre-training over three backbones and two heads that use Mobile-Former \cite{MobileFormer-2022-cvpr} end-to-end in DETR \cite{nicolas2020detr} framework. The backbone is \textit{frozen} for all pre-training methods. QB-Heat significantly outperforms both MoCo-v2 and supervised counterparts. 2.6+ AP gain is achieved for all six combinations of two heads and three backbones. For the lightest model using Mobile-Former-285M as backbone and \texttt{MF-Dec-211} as head, 5.2 AP is gained. Similar trend is observed when evaluating in RetinaNet \cite{Lin_2017_ICCV_retinanet_focal} framework (see \cref{table:retina-results} in \cref{apx:more-exp}). This demonstrates that our QB-Heat learns better spatial representation via quarter-block prediction. 

\vspace{1mm} \noindent
\textbf{QB-Heat and ImageNet-1K fine-tuning provides strong task-agnostic encoders:}
Interestingly, fine-tuning on ImageNet-1K alone (but freezing on COCO) after QB-Heat pre-training introduces consistent gain on object detection. As shown in \cref{table:coco-det-results}, it gains 1.4--4.3 AP over six combinations of three encoders and two detection decoders. \cref{fig:agnostic-enc} plots performances of classification and detection that are achieved by sharing encoder weights (or task-agnostic encoder). Although QB-Heat is far behind ImageNet-1K supervised pre-training on classification, it overtakes by a clear margin in detection, showcasing better spatial representation. Fine-tuning on ImageNet-1K boosts performances of both tasks, providing strong task-agnostic encoders. As fine-tuning is performed with layer-wise learning rate decay, it essentially leverages advantages of both QB-Heat (spatial representation at lower levels) and class supervision (semantic representation at higher levels). 

\vspace{1mm} \noindent
\textbf{Discussion:}
Compared to QB-Heat, we observe two \textit{unexpected} behaviors in MoCo-v2, especially when using larger models (MF-1.0G, MF-3.7G). Firstly, the \texttt{tran-1} probing performance does not improve when using wider decoders (see \cref{fig:in1k-decoder-probing}). Secondly, larger backbones have more degradation in detection performance. As shown in the bottom half of \cref{table:coco-det-results}, models with descending size (MF-3.7G, MF-1.0G, MF-285M) have ascending AP (12.2, 16.9, 22.1). We believe this is because MoCo-v2 focuses more on semantics than spatial representation, providing less room for the following \texttt{tran-1} decoder to improve via spatial fusion. Also, the lack of spatial representation makes it difficult to regress object from sparse queries in DETR. 
Detailed analysis is provided in \cref{apx:analysis-moco}.

\subsection{Fine-tuning on Individual Tasks}
Below, we compare with prior works on fine-tuning results of both classification and detection. End-to-end comparison (combining  architecture and pre-training) is performed and grouped by computational complexity (FLOPs).

\vspace{1mm} \noindent
\textbf{ImageNet-1K classification:}
\cref{table:imgnet-cmp} shows that QB-Heat pre-trained Mobile-Former has comparable performance to ViTs pre-trained by either contrastive or MIM methods. This showcases that proper design (quarter masking and linear prediction) of masked image modeling achieves decent performance for CNN based Mobile-Former. 

\begin{table}[t]
	\begin{center}
	    \smallfootnote
	    \setlength{\tabcolsep}{3.0mm}{
		\begin{tabular}{ll|rr|ccc}
		    \specialrule{.1em}{.05em}{.05em} 
			pre-train & encoder & madds & param & fine-tune \\
			\specialrule{.1em}{.05em}{.05em} 
			MAE-Lite \cite{maelite2022} & ViT-Tiny & 1.2G & 6M& 76.1 \\
			\textbf{QB-Heat} & MF-285M & 0.4G & 6M & 75.8 \\
            \textbf{QB-Heat} & MF-1.0G & 1.4G & 15M & 80.5 \\ 
			\hline
			iBOT \cite{zhou2021ibot} & ViT-S & 4.6G & 22M & 82.3 \\
			MoCo-v3 \cite{chen2021mocov3} & ViT-S & 4.6G & 22M & 81.4 \\
			MAE \cite{MaskedAutoencoders2021} & ViT-S & 4.6G & 22M & 79.5 \\
			CMAE \cite{huang2022cmae} & ViT-S & 4.6G & 22M & 80.2 \\
			ConvMAE \cite{gao2022convmae} & ConvViT-S & 6.4G & 22M & 82.6 \\ 
			\textbf{QB-Heat} & MF-3.7G & 5.5G & 35M& 82.5 \\
            
			\specialrule{.1em}{.05em}{.05em} 
		\end{tabular}
		}
	\end{center}
	\vspace{-4mm}
	\caption{\textbf{Comparisons with previous results on ImageNet-1K.} All self-supervised methods are evaluated by end-to-end fine-tuning. All results are on an image size of 224.}
	\label{table:imgnet-cmp}
	\vspace{-2mm}
\end{table}

\begin{table}[t!]
	\begin{center}
	    \smallfootnote
	    \setlength{\tabcolsep}{0.6mm}{
		\begin{tabular}{l|c|ccc|ccc|c|c}
		    \specialrule{.1em}{.05em}{.05em} 
			\multirow{2}{*}{model} & 
			\multirow{2}{*}{query} &
			\multirow{2}{*}{AP} & \multirow{2}{*}{AP\textsubscript{50}} & \multirow{2}{*}{AP\textsubscript{75}} & \multirow{2}{*}{AP\textsubscript{S}} & \multirow{2}{*}{AP\textsubscript{M}} & \multirow{2}{*}{AP\textsubscript{L}} & madds & param  \\
	         &  &  &  &  &  &  & & (G) & (M)  \\	
			\specialrule{.1em}{.05em}{.05em} 
			
            DETR-DC5\cite{nicolas2020detr} & \textbf{100} &  43.3  & 63.1 & 45.9 & 22.5 & 47.3 & 61.1 & 187 &	41 \\
            Deform-DETR\cite{zhu2020deformable} & 300 &  46.2  & 65.2 & 50.0 & 28.8 & 49.2 & 61.7 & 173 &	\textbf{40} \\
            DAB-DETR\cite{liu2022dabdetr} & 900 & 46.9  & 66.0 & 50.8 & 30.1 & 50.4 & 62.5 & 195 &	48 \\
            DN-DETR\cite{li2022dn} & 900 & 48.6  & 67.4 & 52.7 & 31.0 & 52.0 & 63.7 & 195 &	48 \\
            DINO\cite{zhang2022dino} & 900 & \textbf{50.9}  & \textbf{69.0} & \textbf{55.3} & \textbf{34.6} & \textbf{54.1} & 64.6 & 279 &	47 \\
		    \hline
            \textbf{QB-MF-DETR} & \textbf{100} & 49.0  & 67.8 & 53.4 & 30.0 & 52.8 & \textbf{65.8} & \textbf{112} & 44\\
			\specialrule{.1em}{.05em}{.05em} 
		\end{tabular}
		}
	\end{center}
	\vspace{-4mm}
	\caption{\textbf{Comparisons with DETR based models on COCO}. QB-MF-DETR uses Mobile-Former (MF-3.7G) as backbone, which has similar FLOPs and model size with ResNet-50 used in other methods. MAdds is based on image size 800$\times$1333.}
	\label{table:coco-det-detr-cmp}
	\vspace{-3mm}
\end{table}

\vspace{1mm} \noindent
\textbf{COCO object detection:}
Fine-tuning backbone on COCO further boosts detection performance. \cref{table:coco-det-detr-cmp} shows that QB-MF-DETR (QB-Heat pre-trained Mobile-Former in DETR framework) achieves 49.0 AP, outperforming most of DETR based detectors except DINO \cite{zhang2022dino}. However, our method uses significantly less FLOPs (112G vs. 279G) with significantly fewer object queries (100 vs. 900).    
Full comparison of fine-tuning results over pre-training methods is reported in \cref{table:detr-results-ft} in \cref{apx:more-exp}.

\section{Discussion}
\begin{figure}[t]
	\begin{center}
		\includegraphics[width=1.0\linewidth]{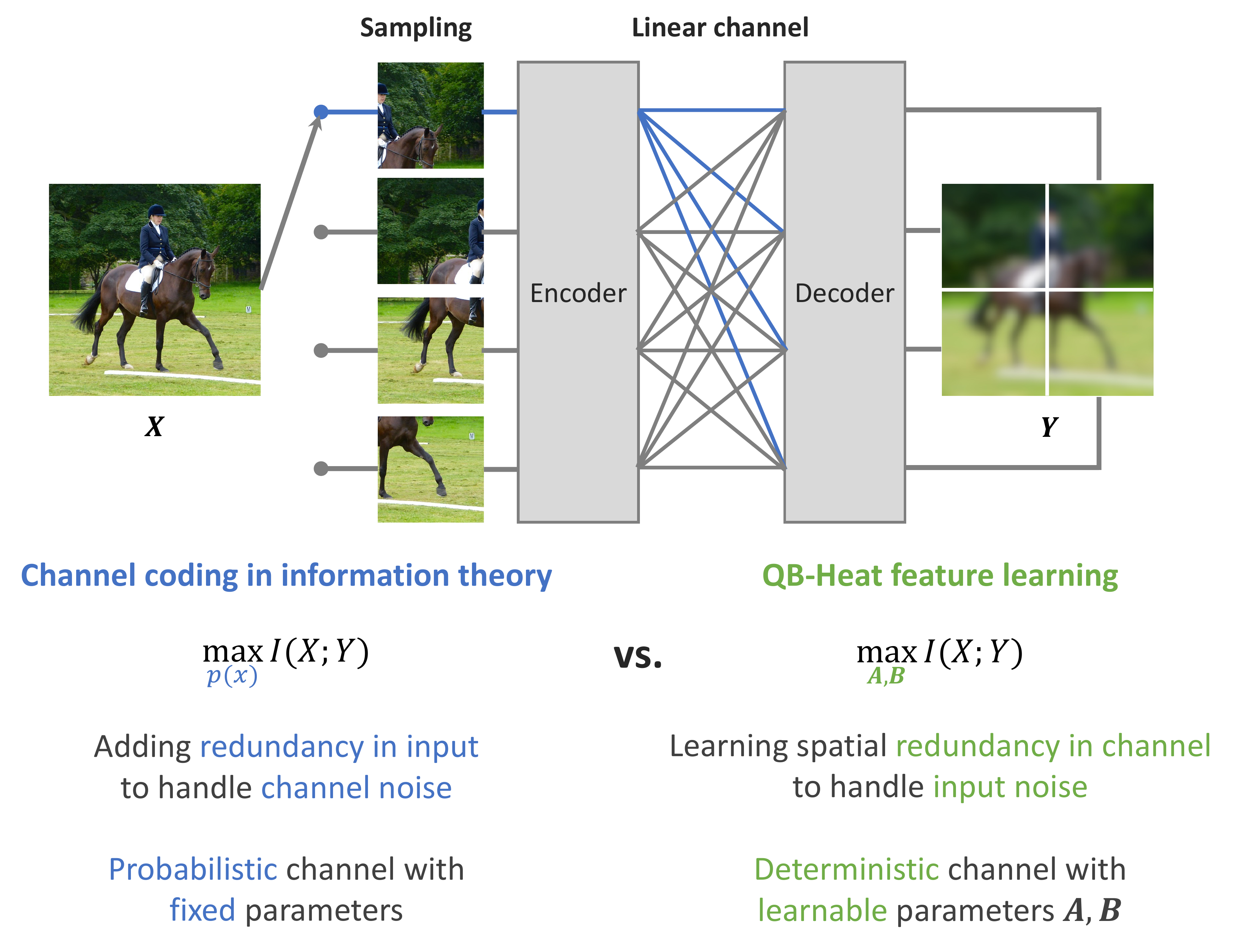}
	\end{center}
	\vspace{-5mm}
	\caption{\textbf{Duality between channel coding in information theory and QB-Heat feature learning.} QB-Heat pre-training can be considered as a communication system with quarter sampling and linear channel. Best viewed in color. }
	\label{fig:comm-system}
	\vspace{-2mm}
\end{figure}

\noindent
\textbf{Connection with information theory:}
Essentially, QB-Heat is a communication system (see \cref{fig:comm-system}) that communicates a quarter of image (via quarter sampling) through linear channels to reconstruct the whole image. It follows the channel capacity in information theory \cite{Cover2006} to maximize the mutual information between input and output, but introduces interesting  differences in channel, input and optimization (see \cref{fig:comm-system}). Compared to information channel coding, where the channel is \textit{probabilistic} with \textit{fixed} parameters (e.g. probability transmission matrix of symmetric channels) and the optimization is over the \textit{input distribution} $p(x)$, QB-Heat has \textit{deterministic} linear channel with \textit{learnable} parameters ($\bm{A}$, $\bm{B}$ in \cref{eq:pde-linear}) to optimize. 

The key insight is the \textbf{\textit{duality of noise handling}} between QB-Heat and information channel coding. The channel coding theorem combats \textit{noise in the channel} by adding redundancy on input $X$ in a controlled fashion, while QB-Heat handles \textit{noisy input} (i.e. corrupted image) by learning feature representation and spatial redundancy jointly.

\vspace{1mm} \noindent
\textbf{Connection with diffusion model:} Compared to diffusion models \cite{NEURIPS2020_ddpm, song2019generative, song2021scorebased} that study \textit{noise diffusion} along the path from image to noise, QB-Heat studies a different type of diffusion: i.e. the \textit{semantic heat diffusion} of feature vector across 2D space.  But they essentially share a common insight: \textbf{\textit{learning diffusion rate as a function of signal.}} Specifically, in diffusion model, the noise $\bm{\epsilon}_t$ at step $t$ is a function of $\bm{x}_t$. In contrast, QB-Heat models the feature change using linear equations as $\frac{\partial \bm{z}}{\partial x}=\bm{A}\bm{z}$, $\frac{\partial \bm{z}}{\partial y}=\bm{B}\bm{z}$.

\vspace{1mm} \noindent
\textbf{Limitations:} 
QB-Heat has a major limitation: not working well for vision transformers (ViT). This is mainly due to the discrepancy between pre-training and inference on the range of token interaction. Specifically, QB-Heat does not have a chance to see tokens beyond a quarter-block in pre-training, but all tokens of an entire image are used during inference. This discrepancy becomes more critical for transformers than CNN, as long range interaction is directly modeled via attention in transformer.

\section{Conclusion}
This paper presents a new self-supervised learning guided by heat equation. It extends heat equation from a single measurable variable to high dimensional latent feature vector, and simplifies it into first order linear differential equations. Based on such simplification, we develop a new masked image modeling (named QB-Heat) that learns to linearly predict three masked quarter-blocks from a single unmasked quarter-block. QB-Heat not only enables masked image modeling for efficient CNN based architectures, but also provides strong task-agnostic encoders on both image classification and object detection. We hope this encourage new understanding of representative feature space by leveraging principles in physics. 

%\clearpage
%%%%%%%%% REFERENCES
{\small
\bibliographystyle{ieee_fullname}
\bibliography{egbib}
}

%%%%%%%%%%%%%%%%%%%%%%%%%%%%%%%%%%%%%%%%%%%%%%%%%%%%%%%%%%%%%%%%%%%%%%%%
\clearpage
\appendix

\section{Derivation of Implicit Linear Models}
\label{app:derivation-linear}
Below, we show how to derive linear models for implicit directions from explicit directions (see \cref{fig:num-lin-models}). Let us denote the two explicit models in \cref{eq:pde-finite} along $x$ and $y$ axes (right and down) as $\bm{A}_1$ and $\bm{B}_1$. Firstly, we derive the models along the negative directions of $x$ and $y$ axes ($\bm{A}_2$ and $\bm{B}_2$). Then we further extend to 4 diagonal directions ($\bm{C}_{11}$, $\bm{C}_{12}$, $\bm{C}_{21}$, $\bm{C}_{22}$). 

\vspace{2mm} \noindent
\textbf{Computing $\bm{A}_2$ and $\bm{B}_2$}: Below, we show how to compute $\bm{A}_2$ from $\bm{A}_1$. $\bm{B}_2$ can be derived similarly. The finite difference approximations along opposite directions by using $\bm{A}_1$ and $\bm{A}_2$ are represented as:
\begin{align}
\bm{z}(x, y)&=(\bm{I}+\Delta x\bm{A}_1)\bm{z}(x-\Delta x,y) \nonumber \\
\bm{z}(x-\Delta x, y)&=(\bm{I}+\Delta x\bm{A}_2)\bm{z}(x,y).
\label{eq:lin-derive-1}
\end{align}
Thus, $\bm{I}$$+$$\Delta x\bm{A}_2$ is the inverse matrix of $\bm{I}$$+$$\Delta x\bm{A}_1$. To avoid the difficulty of computing inverse, we approximate $\bm{A}_2$ as follows:
\begin{align}
\bm{A}_2&=\frac{(\bm{I}+\Delta x\bm{A}_1)^{-1}-\bm{I}}{\Delta x} \approx  -\bm{A}_1.
\label{eq:lin-derive-2}
\end{align}
This is used when only two explicit linear models $\bm{A}_1$ and $\bm{B}_1$ are available (see \cref{fig:num-lin-models}). 

\vspace{2mm} \noindent
\textbf{Computing $\bm{C}_{11}$, $\bm{C}_{12}$, $\bm{C}_{21}$, $\bm{C}_{22}$}: We show the derivation of $\bm{C}_{11}$ from $\bm{A}_1$ and $\bm{B}_1$. The other three diagonal directions can be derived similarly. The diagonal prediction of $\bm{z}(x+\Delta x, y+\Delta y)$ from $\bm{z}(x, y)$ can be achieved in two steps (i.e. horizontal prediction followed by vertical) as:
\begin{align}
\bm{z}(x+\Delta x, y+\Delta y)&=(\bm{I}+\sqrt{\Delta x^2+\Delta y^2}\bm{C}_{11})\bm{z}(x,y) \nonumber \\
&=(\bm{I}+\Delta x\bm{A}_1)\bm{z}(x,y+\Delta y) \nonumber \\
&=(\bm{I}+\Delta x\bm{A}_1)(\bm{I}+\Delta y\bm{B}_1)\bm{z}(x,y).
\label{eq:lin-derive-3}
\end{align}
The order of horizontal and vertical difference can be flipped as:
\begin{align}
\bm{z}(x+\Delta x, y+\Delta y)&=(\bm{I}+\Delta y\bm{B}_1)\bm{z}(x+\Delta x,y) \nonumber \\
&=(\bm{I}+\Delta y\bm{B}_1)(\bm{I}+\Delta x\bm{A}_1)\bm{z}(x,y). 
\label{eq:lin-derive-4}
\end{align}
Given that Eq. \ref{eq:lin-derive-3} and Eq. \ref{eq:lin-derive-4} are identical, $\bm{A}_1$ and $\bm{B}_1$ commute (i.e. $\bm{A}_1\bm{B}_1=\bm{B}_1\bm{A}_1$). In practical, $\bm{C}_{11}$ is computed by averaging Eq. \ref{eq:lin-derive-3} and Eq. \ref{eq:lin-derive-4} as follows:
\begin{align}
\bm{C}_{11}&=\frac{\Delta x\bm{A}_1+\Delta y\bm{B}_1+\Delta x \Delta y(\bm{A}_1\bm{B}_1+\bm{B}_1\bm{A}_1)/2}{\sqrt{\Delta x^2+\Delta y^2}}.
\label{eq:lin-derive-5}
\end{align}
Note this computation is needed when using 2 or 4 explicit linear models. (see \cref{fig:num-lin-models}).

\iffalse
\begin{figure}[t]
	\begin{center}
		\includegraphics[width=1.0\linewidth]{figures/linear-num.pdf}
	\end{center}
	\vspace{-4mm}
	\caption{\textbf{Number of explicit linear models.} A solid arrow indicates an explicit linear model, while a dash arrow indicates an implicit model derived from the explicit counterparts.}
	\label{fig:num-lin-models-supp}
	\vspace{-3mm}
\end{figure}
\fi

%%%%%%%%%%%%%%%%%%%%%%%%%%%%%%%%%%%%%%%%%%%%%%%%%%%%%%%%%%%%%%
\section{Implementation Details} \label{apx:arch-all}
\subsection{ImageNet Experiments} \label{apx:arch-imgnet}
\vspace{1mm} \noindent
\textbf{Mobile-Former encoders:}
\cref{table:mf-enc-achs} shows the network details for three variants of Mobile-Former \cite{MobileFormer-2022-cvpr} used in this paper. All of them have 12 blocks and 6 global tokens, but different widths. They are used as encoder (or backbone) for both image classification and object detection. Note that they only have 4 stages and output at resolution ($\frac{1}{16}$), providing more spatial details for translational prediction. These models are manually designed without searching for the optimal architecture parameters (e.g. width or depth).  

\begin{table}[t]
	\begin{center}
	    \footnotesize
	    \setlength{\tabcolsep}{1.0mm}{
		\begin{tabular}{c|c|c|cc|cc|cc}
		    \specialrule{.1em}{.05em}{.05em} 
			\multirow{2}{*}{stage}&
			\multirow{2}{*}{resolution}&
			\multirow{2}{*}{block}&
			\multicolumn{2}{c|}{\textbf{MF-3.7G}} & \multicolumn{2}{c|}{\textbf{MF-1.0G}} & \multicolumn{2}{c}{\textbf{MF-285M}} \\
			\cline{4-9}
			 & & & \#exp & \#out & \#exp & \#out &  \#exp & \#out  \\
		
			\specialrule{.1em}{.05em}{.05em} 
			token & & & \multicolumn{2}{c|}{6$\times$256} & \multicolumn{2}{c|}{6$\times$256} & \multicolumn{2}{c}{6$\times$192}   \\
			\hline
			stem & 224$^2$ & conv 3$\times$3 & -- & 64 &  -- & 32 &   -- & 16  \\
			
			\hline
			1 & 112$^2$ & bneck-lite & 128 & 64 &   64 & 32 &   32 & 16   \\
			\hline
			\multirow{2}{*}{2} & \multirow{2}{*}{56$^2$} & M-F$^{\downarrow}$ & 384 & 112 & 192 & 56 &  96 & 28 \\
			& &  M-F & 336 & 112 & 168 & 56 & 84 & 28  \\
			\hline
			\multirow{3}{*}{3} & \multirow{3}{*}{28$^2$} &  M-F$^{\downarrow}$ & 672 & 192 &  336 & 96 &  168 & 48  \\
			& &  M-F  & 576 & 192 & 288 & 96 & 144 & 48  \\
			& &  M-F  & 576 & 192 & 288 & 96 & 144 & 48  \\
			\hline
			\multirow{7}{*}{4} & \multirow{7}{*}{14$^2$} &  M-F$^{\downarrow}$ & 1152 & 352 &  288 & 96 &  240 & 80  \\
			& &  M-F & 1408 & 352 &  704 & 176 &  320 & 88  \\
			& &  M-F & 1408 & 352 &  704 & 176 &  480 & 88  \\
			& &  M-F & 2112 & 480 &  1056 & 240 & 528 & 120   \\
			& &  M-F & 2880 & 480 &  1440 & 240 & 720 & 120   \\
			& &  M-F & 2880 & 480 &. 1440 & 240 & 720 & 120   \\
			& & conv 1$\times$1 & -- & 2880 &   -- & 1440 & -- & 720  \\
			
			\specialrule{.1em}{.05em}{.05em} 
			pool & \multirow{2}{*}{1$^2$} & \multirow{2}{*}{--} & \multirow{2}{*}{--} & \multirow{2}{*}{3136} &  \multirow{2}{*}{--} & \multirow{2}{*}{1696} &  \multirow{2}{*}{--} & \multirow{2}{*}{912}  \\
			concat & &  &  &  &  &   &  &  \\
			\specialrule{.1em}{.05em}{.05em} 
		\end{tabular}
		}
	\end{center}
	\vspace{-1mm}
	\caption{\textbf{Specification of Mobile-Former encoders}. ``bneck-lite" denotes the lite bottleneck block \cite{li2021micronet}. ``M-F" denotes the Mobile-Former block and ``M-F$^{\downarrow}$" denotes the Mobile-Former block for downsampling.}
	\vspace{-1mm}
	\label{table:mf-enc-achs}
\end{table}

\vspace{1mm} \noindent
\textbf{QB-Heat pre-training setup:} 
\cref{table:pretrain-setting} shows the pre-training setting. The learning rate is scaled as \textit{lr = base\_lr}$\times$batchsize / 256. We use image size 256 such that the output feature resolution is multiple of 4 (i.e. 16$\times$16). This is required for prediction from the unmasked quarter-block at center position.

\vspace{1mm} \noindent
\textbf{Linear probing:} Our linear probing follows \cite{MaskedAutoencoders2021} to adopt an extra BatchNorm layer without
affine transformation (affine=False). See detailed setting in  \cref{table:lin-setting}.

\vspace{1mm} \noindent
\textbf{\texttt{tran-1} probing:} \cref{table:tran1-setting} shows the setting for \texttt{tran-1} decoder probing. Note that the default decoder widths are 192, 384, 768 for MF-285M, MF-1.0G and MF-3.7G, respectively.

\vspace{1mm} \noindent
\textbf{End-to-end fine-tuning:} \cref{table:finetune-setting} shows the setting for end-to-end fine-tuning of both encoder and \texttt{tran-1} decoder. The decoder weights are initialized from \texttt{tran-1} probing.

\begin{table}%[t!]
	\begin{center}
	    \footnotesize
	    \setlength{\tabcolsep}{8.3mm}{
		\begin{tabular}{l|l }
		    \specialrule{.1em}{.05em}{.05em} 
			config & value   \\
			\hline
			optimizer & AdamW \\
			base learning rate & 1.5e-4 \\
			weight decay & 0.1 \\
			batch size & 1024 \\
			learning rate schedule & cosine decay \\
			warmup epochs & 10 \\
			image size & 256$^2$ \\
			augmentation & RandomResizeCrop \\
			\specialrule{.1em}{.05em}{.05em} 
		\end{tabular}
		}
	\end{center}
	\vspace{-2mm}
	\caption{\textbf{Pre-training setting.}}
	\label{table:pretrain-setting}
\end{table}

\begin{table}%[t!]
	\begin{center}
	    \footnotesize
	    \setlength{\tabcolsep}{8.3mm}{
		\begin{tabular}{l|l }
		    \specialrule{.1em}{.05em}{.05em} 
			config & value   \\
			\hline
			optimizer & SGD \\
			base learning rate & 0.1 \\
			weight decay & 0 \\
			batch size & 4096 \\
			learning rate schedule & cosine decay \\
			warmup epochs & 10 \\
			training epochs & 90 \\
			augmentation & RandomResizeCrop \\
			\specialrule{.1em}{.05em}{.05em} 
		\end{tabular}
		}
	\end{center}
	\vspace{-2mm}
	\caption{\textbf{Linear probing setting.}}
	\label{table:lin-setting}
\end{table}

\begin{table}%[t!]
	\begin{center}
	    \footnotesize
	    \setlength{\tabcolsep}{3.8mm}{
		\begin{tabular}{l|l }
		    \specialrule{.1em}{.05em}{.05em} 
			config & value   \\
			\hline
			
			optimizer & AdamW \\
			base learning rate & 0.0005 \\
			weight decay & 0.1 \\
			batch size & 4096 \\
			learning rate schedule & cosine decay \\
			warmup epochs & 10 \\
			training epochs & 200 \\
			augmentation & RandAug (9, 0.5) \\
			label smoothing & 0.1 \\
			dropout & 0.1 (MF-285M) 0.2 (MF-1.0G/3.7G) \\
			random erase & 0 (MF-285M/1.0G) 0.25 (MF-3.7G) \\
			\specialrule{.1em}{.05em}{.05em} 
		\end{tabular}
		}
	\end{center}
	\vspace{-2mm}
	\caption{\textbf{\texttt{tran-1} probing setting.}}
	\label{table:tran1-setting}
\end{table}

\begin{table}%[t!]
	\begin{center}
	    \footnotesize
	    \setlength{\tabcolsep}{0.6mm}{
		\begin{tabular}{l|l }
		    \specialrule{.1em}{.05em}{.05em} 
			config & value   \\
			\hline
			optimizer & AdamW \\
			base learning rate & 0.0005 \\
			weight decay & 0.05 \\
			layer-wise lr decay & 0.90 (MF-285M/1.0G) 0.85 (MF-3.7G)\\
			batch size & 512 \\
			learning rate schedule & cosine decay \\
			warmup epochs & 5 \\
			training epochs & 200 (MF-285M) 150 (MF-1.0G) 100 (MF-3.7G) \\
			augmentation & RandAug (9, 0.5) \\
			label smoothing & 0.1 \\
			mixup & 0 (MF-285M) 0.2 (MF-1.0G) 0.8 (MF-3.7G) \\
			cutmix & 0 (MF-285M) 0.25 (MF-1.0G) 1.0 (MF-3.7G) \\
			dropout & 0.2 \\
			random erase & 0.25 \\
			\specialrule{.1em}{.05em}{.05em} 
		\end{tabular}
		}
	\end{center}
	\vspace{-2mm}
	\caption{\textbf{End-to-end fine-tuning setting.}}
	\label{table:finetune-setting}
\end{table}

\subsection{Object Detection in COCO} \label{apx:arch-coco}
\vspace{1mm} \noindent
\textbf{MF-DETR decoders for object detection:}
\cref{table:od-mf-dec-achs} shows the two decoder structures that use Mobile-Former \cite{MobileFormer-2022-cvpr} in DETR \cite{nicolas2020detr} framework. Both have 100 object queries with dimension 256. They share similar structure over three scales but have different depths. As the backbone ends at resolution $\frac{1}{16}$, we first perform downsampling (to $\frac{1}{32}$) in the decoder. 

\begin{table}%[b!]
	\begin{center}
	    \footnotesize
	    \setlength{\tabcolsep}{4.3mm}{
		\begin{tabular}{c|cc|cc}
		    \specialrule{.1em}{.05em}{.05em} 
			stage & \multicolumn{2}{c|}{\texttt{MF-Dec-522}} & \multicolumn{2}{c}{\texttt{MF-Dec-211}} \\
		
			\specialrule{.1em}{.05em}{.05em} 
			query & \multicolumn{2}{c|}{100$\times$256} & \multicolumn{2}{c}{100$\times$256}  \\
			\hline
			\multirow{2}{*}{$\frac{1}{32}$} & down-conv & & down-conv & \\ 
			& M-F$^+$ & $\times$5 & M-F$^{+}$ & $\times$2  \\
		
			\hline
			\multirow{2}{*}{$\frac{1}{16}$} & up-conv & & up-conv & \\
			 & M-F$^{-}$ & $\times$2 & M-F$^{-}$ & $\times$1  \\
		    \hline
		    \multirow{2}{*}{$\frac{1}{8}$} & up-conv & &up-conv  & \\	
			 & M-F$^{-}$ &$\times$2 & M-F$^{-}$ &$\times$1 \\	
			\specialrule{.1em}{.05em}{.05em} 
		\end{tabular}
		}
	\end{center}
	\vspace{-1mm}
	\caption{\textbf{Specification of Mobile-Former decoders in COCO object detection}. 100 object queries with dimension 256 are used. 
	``down-conv" denotes a downsampling convolutional block that includes a 3$\times$3 depthwise (stride=2) and a pointwise convolution (256 output channels). ``up-conv" denotes a upsampling convolutional block that includes bilinear interpolation followed by a 3$\times$3 depthwise and a pointwise convolution. ``M-F$^{+}$" and ``M-F$^{-}$" modify the \textit{Mobile} sub-block in the original Mobile-Former block. The former replaces it with a transformer block, while the latter uses lite bottleneck \cite{li2021micronet}.
	} 
	\vspace{-1mm}
	\label{table:od-mf-dec-achs}
\end{table}

\vspace{1mm}
\noindent \textbf{DETR training setup:} In decoder probing with frozen bacbkone, only decoders are trained for 500 epochs on 8 GPUs with 2 images per GPU. AdamW optimizer is used with initial learning rate 1e-4. The learning rate drops by a factor of 10 after 400 epochs. The weight decay is 1e-4 and dropout rate is 0.1. Fine-tuning involves additional 200 epochs from decoder probing. Both encoder and decoder are fine-tuned with initial learning rate 1e-5 which drops to 1e-6 after 150 epochs.

%%%%%%%%%%%%%%%%%%%%%%%%%%%%%%%%%%%%%%%%%%%%%%%%%%%%%%%%%%%%%%
\section{More Experimental Results} \label{apx:more-exp}
\begin{figure*}[t]
	\begin{center}
		\includegraphics[width=1.0\linewidth]{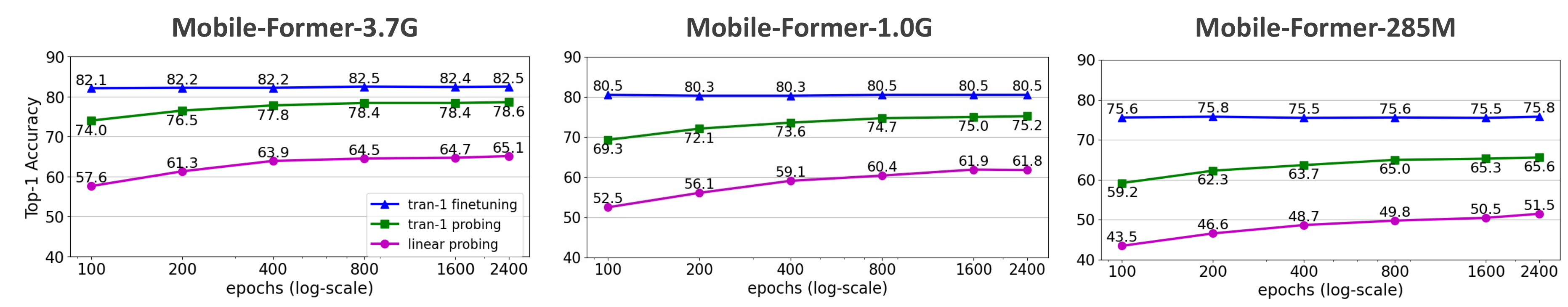}
	\end{center}
	\vspace{-4mm}
	\caption{\textbf{Training schedules.} Longer training schedule provides consistent improvement for linear and \texttt{tran-1} probing over different models, while fine-tuning performance is not sensitive to training schedule. }
	\label{fig:linprob-epochs-all}
\end{figure*}

\noindent
\textbf{Ablation on training schedule:}
\cref{fig:linprob-epochs-all} shows the influence of the length of training schedule for three Mobile-Former encoders. They share the similar trend: the accuracies of two decoder probings (linear and \texttt{tran-1}) improve steadily as training lasts longer, while fine-tuning with \texttt{tran-1} achieves decent performance even for pre-training 100 epochs. This is different from MAE \cite{MaskedAutoencoders2021}, in which fine-tuning relies on longer training to improve. 

\vspace{1mm} \noindent
\textbf{Decoder probing (\textit{frozen backbone}) on COCO object detection in RetinaNet framework:}
\cref{table:retina-results} compares QB-Heat with MoCo-V2 and ImageNet supervised pre-training over three backbones in RetinaNet \cite{Lin_2017_ICCV_retinanet_focal} framework. The backbone is \textit{frozen} for all pre-training methods. Similar to the results in DETR \cite{nicolas2020detr} framework (see \cref{sec:multitask-decoder-probing-results}),  QB-Heat outperforms both MoCo-v2 and supervised counterparts, demonstrating that our QB-Heat learns better spatial representation via quarter-block prediction. In addition, the followed fine-tuning on ImageNet-1K provides consistent gain. 

\begin{table*}[t]
	\begin{center}
	    \smallfootnote
	    \setlength{\tabcolsep}{2.5mm}{
		\begin{tabular}{ccc|ccccc|lcc|ccc}
		    \specialrule{.1em}{.05em}{.05em} 
			& \textbf{head} & &
			\multicolumn{5}{c|}{\textbf{backbone}}&
			\multirow{3}{*}{AP} & 
			\multirow{3}{*}{AP\textsubscript{50}} & \multirow{3}{*}{AP\textsubscript{75}} & \multirow{3}{*}{AP\textsubscript{S}} & \multirow{3}{*}{AP\textsubscript{M}} & \multirow{3}{*}{AP\textsubscript{L}} \\
	         model & madds & param & model & madds & param & pre-train & IN1K-ft & &  & & &   \\
	         & (G) & (M) & & (G) & (M) & & & & & & & \\
	         
			\specialrule{.1em}{.05em}{.05em} 
			\multirow{12}{*}{RetinaNet} & \multirow{4}{*}{237.7} & \multirow{4}{*}{61.4} & \multirow{4}{*}{\makecell{MF-3.7G}} & \multirow{4}{*}{77.5} & \multirow{4}{*}{25.0} & sup & -- & 34.0 & 54.4 & 35.3 & 18.7 & 36.0 & 46.1\\
			& & & & & & moco2 & \xmark     & 29.4\textsubscript{\textcolor{red}{(-4.6)}} & 47.8 & 30.7 & 20.5 & 30.7 & 35.1 \\
			& & & & & & \textbf{QB-Heat} & \xmark & 36.6\textsubscript{\color{ForestGreen}{(+2.6)}} & 55.8 & 38.6 & 20.8 & 39.9 & 47.7 \\
			& & & & & & \textbf{QB-Heat} & \checkmark & \textbf{38.7} \textsubscript{\color{ForestGreen}{(+4.7)}}& \textbf{59.0} & \textbf{41.0} & \textbf{23.0} & \textbf{42.0} & \textbf{49.9} \\
			\cline{2-14}
			& \multirow{4}{*}{178.1} & \multirow{4}{*}{17.2} & \multirow{4}{*}{\makecell{MF-1.0G}} 
            & \multirow{4}{*}{20.4} & \multirow{4}{*}{11.7} & sup & --& 33.6 & 54.0 & 34.9 & 20.9 & 35.8 & 44.4 \\
			& & & & & & moco2 & \xmark & 29.3\textsubscript{\textcolor{red}{(-4.3)}} & 47.4 & 30.4 & 17.6 & 30.5 & 37.6 \\
			& & & & & & \textbf{QB-Heat} & \xmark & 35.7\textsubscript{\color{ForestGreen}{(+2.1)}} & 54.7 & 37.8 & 20.5 & 38.8 & 46.8  \\
			& & & & & & \textbf{QB-Heat} & \checkmark & \textbf{37.7}\textsubscript{\color{ForestGreen}{(+4.1)}} & \textbf{57.8}& \textbf{40.0} & \textbf{22.7} &\textbf{40.6} & \textbf{48.8} \\
			\cline{2-14}
			& \multirow{4}{*}{163.1} & \multirow{4}{*}{9.5} & \multirow{4}{*}{\makecell{MF-285M}} 
            & \multirow{4}{*}{5.6} & \multirow{4}{*}{4.9} & sup & --& 30.8 & 50.0 & 31.9 & 17.5 & 32.4 & 41.4 \\
			& & & & & & moco2 & \xmark & 28.0\textsubscript{\textcolor{red}{(-2.8)}} & 45.7 & 29.2 & 16.1 & 29.1 & 37.7 \\
			& & & & & & \textbf{QB-Heat} & \xmark & 31.3\textsubscript{\color{ForestGreen}{(+0.5)}} & 49.2 & 33.1 & 18.1 & 33.3 & 42.4 \\
			& & & & & & \textbf{QB-Heat} & \checkmark & \textbf{33.8}\textsubscript{\color{ForestGreen}{(+3.0)}} & \textbf{52.6} & \textbf{35.5} & \textbf{19.7} & \textbf{36.4} & \textbf{44.5} \\
			\specialrule{.1em}{.05em}{.05em} 
		\end{tabular}
		}
	\end{center}
	\vspace{-3mm}
	\caption{\textbf{COCO object detection results} on \texttt{val2017} for \textbf{\textit{frozen}} backbone pre-trained on ImageNet-1K. Evaluation is conducted over three backbones in RetinaNet \cite{nicolas2020detr} framework. Our QB-Heat outperforms MoCo-v2 and supervised baselines. Fine-tuning on ImageNet-1K provides consistent improvement. Initial ``MF" (e.g. MF-3.7G) refers to Mobile-Former. ``IN1K-ft" indicates fine-tuning on ImageNet-1K. MAdds is based on the image size 800$\times$1333.}
	\label{table:retina-results}
	\vspace{-2mm}
\end{table*}

\vspace{1mm} \noindent 
\textbf{Fine-tuning on ImageNet classification with deeper decoders:}
In \cref{sec:multitask-decoder-probing-results}, we show the results (see \cref{table:qbheat-vs-sup}) for shallow decoders in image classification (linear \texttt{lin} and single transformer block \texttt{tran-1}). We also find that performance can be further improved by adding more transformer blocks (deeper decoder). \cref{table:deeper-enc} shows the fine-tuning results for using 4 transformer blocks \texttt{tran-4}. Compared with MAE \cite{MaskedAutoencoders2021} and MoCo \cite{chen2021mocov3} on ViT \cite{dosovitskiy2021vit}, our QB-Heat achieves either similar results with lower FLOPs and fewer parameters, or better performance with similar FLOPs and number of parameters.

\begin{table}[t]
	\begin{center}
	    \smallfootnote
	    \setlength{\tabcolsep}{1.2mm}{
		\begin{tabular}{lll|rr|ccc}
		    \specialrule{.1em}{.05em}{.05em} 
			pre-train & encoder & decoder & madds & param & fine-tune \\
			\specialrule{.1em}{.05em}{.05em} 
			MAE-Lite \cite{maelite2022} & ViT-Tiny & \texttt{lin} & 1.2G & 6M& 76.1 \\
			\textbf{QB-Heat} & MF-285M & \texttt{tran-4} (d192) & 0.7G & 7M & 78.4 \\
			\hline
			MoCo-v3 \cite{chen2021mocov3} & ViT-S & \texttt{lin} & 4.6G & 22M & 81.4 \\
			MAE \cite{MaskedAutoencoders2021} & ViT-S & \texttt{lin} & 4.6G & 22M & 79.5 \\
			\textbf{QB-Heat} & MF-1.0G & \texttt{tran-4} (d384) & 2.6G & 20M & 81.9 \\
			
			\hline
			MoCo-v3 \cite{chen2021mocov3} & ViT-B & \texttt{lin} & 16.8G & 86M & 83.2 \\
			MAE \cite{MaskedAutoencoders2021} & ViT-B & \texttt{lin} & 16.8G & 86M & 83.6 \\
			\textbf{QB-Heat} & MF-3.7G & \texttt{tran-4} (d768) & 9.9G & 57M& 83.5 \\
            
			\specialrule{.1em}{.05em}{.05em} 
		\end{tabular}
		}
	\end{center}
	\vspace{-4mm}
	\caption{\textbf{Fine-tuning on ImageNet-1K \cite{deng2009imagenet} with deeper decoders.} ``\texttt{tran-4} (d192)" denotes the decoder including 4 transformer blocks with 192 channels. All methods are evaluated by end-to-end fine-tuning. All results are on an image size of 224.}
	\label{table:deeper-enc}
	\vspace{-2mm}
\end{table}

\vspace{1mm} \noindent
\textbf{Fine-tuning on COCO object detection in DETR framework:}  Fine-tuning backbone on COCO further boosts detection performance. \cref{table:detr-results-ft} shows the full comparison of fine-tuning results that use Mobile-Former \cite{MobileFormer-2022-cvpr} end-to-end in DETR \cite{nicolas2020detr} framework. Similar to decoder probing with frozen backbone (see \cref{table:coco-det-results} in \cref{sec:multitask-decoder-probing-results}), QB-Heat clearly outperforms MoCo-v2. But different with decoder probing with frozen backbone where QB-Heat outperforms supervised pre-training by a clear margin (see \cref{table:coco-det-results}), they are on par in COCO fine-tuning. This is because the advantage of QB-Heat pre-training on spatial representation diminishes as the object labels in COCO provide strong guidance. But QB-Heat can still hold its leading position by leveraging fine-tuning on ImageNet-1K to improve semantic representation and transfer it to object detection. As shown in \cref{table:detr-results-ft}, compared to supervised pre-training on ImageNet-1K, QB-Heat pre-training followed by ImageNet-1K fine-tuning gains 0.9--2.0 AP over the supervised baseline for all three backbones and two heads.

\begin{table*}[t]
	\begin{center}
	    \smallfootnote
	    \setlength{\tabcolsep}{2.3mm}{
		\begin{tabular}{ccc|ccccc|lcc|ccc}
		    \specialrule{.1em}{.05em}{.05em} 
			& \textbf{head} & &
			\multicolumn{5}{c|}{\textbf{backbone}}&
			\multirow{3}{*}{AP} & 
			\multirow{3}{*}{AP\textsubscript{50}} & \multirow{3}{*}{AP\textsubscript{75}} & \multirow{3}{*}{AP\textsubscript{S}} & \multirow{3}{*}{AP\textsubscript{M}} & \multirow{3}{*}{AP\textsubscript{L}} \\
	         model & madds & param & model & madds & param & pre-train & IN1K-ft & &  & & &   \\
	         & (G) & (M) & & (G) & (M) & & & & & & & \\
	         
			\specialrule{.1em}{.05em}{.05em} 
			\multirow{12}{*}{\makecell{\texttt{MF-Dec-522}}} & \multirow{4}{*}{34.6} & \multirow{4}{*}{19.4} & \multirow{4}{*}{\makecell{MF-3.7G}} & \multirow{4}{*}{77.5} & \multirow{4}{*}{25.0} & sup & -- & 48.1 & 66.6 & 52.5 & 29.7 & 51.8 & 64.0\\
			& & & & & & moco2 & \xmark     & 41.1\textsubscript{\textcolor{red}{(-7.0)}} & 59.7 & 44.6 & 24.1 & 44.1 & 55.5 \\
			& & & & & & \textbf{QB-Heat} & \xmark & 48.0\textsubscript{\textcolor{red}{(-0.1)}} & 66.3 & 52.3 & 28.1 & 51.7 & 64.3 \\
			& & & & & & \textbf{QB-Heat} & \checkmark & \textbf{49.0} \textsubscript{\color{ForestGreen}{(+0.9)}}& \textbf{67.8} & \textbf{53.4} & \textbf{30.0} & \textbf{52.8} & \textbf{65.8} \\
			\cline{2-14}
			& \multirow{4}{*}{32.3} & \multirow{4}{*}{18.6} & \multirow{4}{*}{\makecell{MF-1.0G}} 
            & \multirow{4}{*}{20.4} & \multirow{4}{*}{11.7} & sup & --& 46.2 & 64.4 & 50.1 & 27.1 & 49.8 & 62.4 \\
			& & & & & & moco2 & \xmark & 41.7\textsubscript{\textcolor{red}{(-4.5)}} & 59.8 & 45.1 & 24.4 & 44.7 & 55.9 \\
			& & & & & & \textbf{QB-Heat} & \xmark & 46.7\textsubscript{\color{ForestGreen}{(+0.5)}} & 64.9 & 50.8 & 26.3 & 50.5 & 63.4  \\
			& & & & & & \textbf{QB-Heat} & \checkmark & \textbf{47.1}\textsubscript{\color{ForestGreen}{(+0.9)}} & \textbf{65.4}& \textbf{51.2} & \textbf{27.5} &\textbf{50.6} & \textbf{63.9} \\
			\cline{2-14}
			& \multirow{4}{*}{31.1} & \multirow{4}{*}{18.2} & \multirow{4}{*}{\makecell{MF-285M}} 
            & \multirow{4}{*}{5.6} & \multirow{4}{*}{4.9} & sup & --& 42.5 & 60.4 & 46.0 & 23.9 & 46.0 & 58.5 \\
			& & & & & & moco2 & \xmark & 39.6\textsubscript{\textcolor{red}{(-2.9)}} & 57.1 & 42.8 & 20.9 & 42.2 & 55.5 \\
			& & & & & & \textbf{QB-Heat} & \xmark & 42.6\textsubscript{\color{ForestGreen}{(+0.1)}} & 60.8 & 46.2 & 22.8 & 46.2 & 59.3 \\
			& & & & & & \textbf{QB-Heat} & \checkmark & \textbf{44.4}\textsubscript{\color{ForestGreen}{(+1.9)}} & \textbf{62.2} & \textbf{48.1} & \textbf{24.6} & \textbf{47.9} & \textbf{61.7} \\
			\specialrule{.1em}{.05em}{.05em}
			\multirow{12}{*}{\makecell{\texttt{MF-Dec-211}}} & \multirow{4}{*}{15.7} & \multirow{4}{*}{9.2} & \multirow{4}{*}{\makecell{MF-3.7G}} & \multirow{4}{*}{77.5} & \multirow{4}{*}{25.0} & sup & -- & 44.0 & 62.8 & 47.7 & \textbf{25.8} & 47.3 & 60.7 \\
			& & & & & & moco2 & \xmark & 35.9\textsubscript{\textcolor{red}{(-8.1)}} & 54.0 & 38.5 & 19.1 & 38.8 & 48.5 \\
			& & & & & & \textbf{QB-Heat} & \xmark & 44.3\textsubscript{\color{ForestGreen}{(+0.3)}} & 62.5 & 48.1 & 24.9 & 47.5 & 60.8 \\
			& & & & & & \textbf{QB-Heat} & \checkmark & \textbf{46.0}\textsubscript{\color{ForestGreen}{(+2.0)}} & \textbf{64.5} & \textbf{49.9} & 25.5 & \textbf{50.1} & \textbf{62.6} \\
			\cline{2-14}
			& \multirow{4}{*}{13.4} & \multirow{4}{*}{8.4} & \multirow{4}{*}{\makecell{MF-1.0G}} 
            & \multirow{4}{*}{20.4} & \multirow{4}{*}{11.7} & sup & -- & 42.5 & 60.6 & 46.0 & 23.6 & 45.9 & 57.9 \\
			& & & & & & moco2 & \xmark & 33.6\textsubscript{\textcolor{red}{(-8.9)}} & 50.4 & 36.2 & 17.2 & 36.2 & 46.3 \\
			& & & & & & \textbf{QB-Heat} & \xmark & 42.4\textsubscript{\textcolor{red}{(-0.1)}} & 60.0 & 46.0 & 22.0 & 45.6 & 59.8 \\
			& & & & & & \textbf{QB-Heat} & \checkmark & \textbf{43.8}\textsubscript{\color{ForestGreen}{(+1.3)}} & \textbf{61.7} & \textbf{47.4} & \textbf{23.7} & \textbf{47.0} & \textbf{60.9} \\
			\cline{2-14}
			& \multirow{4}{*}{12.2} & \multirow{4}{*}{8.0} & \multirow{4}{*}{\makecell{MF-285M}} 
            & \multirow{4}{*}{5.6} & \multirow{4}{*}{4.9} & sup & -- & 37.6 & 55.1 & 40.4 & 18.9 & 40.6 & 53.8 \\
			& & & & & & moco2 & \xmark & 32.3\textsubscript{\textcolor{red}{(-5.3)}} & 48.2 & 34.5 & 15.4 & 34.3 & 46.1\\
			& & & & & & \textbf{QB-Heat} & \xmark & 37.2 \textsubscript{\textcolor{red}{(-0.4)}} & 54.2 & 39.9 & 18.4 & 39.7 & 53.5\\
			& & & & & & \textbf{QB-Heat} & \checkmark & \textbf{39.3}\textsubscript{\color{ForestGreen}{(+1.7)}} & \textbf{56.8} & \textbf{42.3} & \textbf{19.3} & \textbf{42.0} & \textbf{56.6} \\
			\specialrule{.1em}{.05em}{.05em} 
		\end{tabular}
		}
	\end{center}
	\vspace{-3mm}
	\caption{\textbf{COCO object detection results} on \texttt{val2017} for \textbf{\textit{fine-tuning}} backbone. Evaluation is conducted over three backbones and two heads that use Mobile-Former \cite{MobileFormer-2022-cvpr} end-to-end in DETR \cite{nicolas2020detr} framework. Without using labels in ImageNet-1K, our QB-Heat outperforms MoCo-v2 by a clear margin.  When labels in ImageNet-1K is available, QB-Heat pre-training followed by ImageNet-1K fine-tuning outperforms the supervised baselines. Initial ``MF" (e.g. \texttt{MF-Dec-522}) refers to Mobile-Former. ``IN1K-ft" indicates fine-tuning on ImageNet-1K. MAdds is based on the image size 800$\times$1333.}
	\label{table:detr-results-ft}
	\vspace{-2mm}
\end{table*}

%%%%%%%%%%%%%%%%%%%%%%%%%%%%%%%%%%%%%%%%%%%%%%%%%%%%%%%%%%%%%%
\section{Analysis of Models Pre-trained by MoCo-v2} \label{apx:analysis-moco}

\begin{figure}[t]
	\begin{center}
		\includegraphics[width=1.0\linewidth]{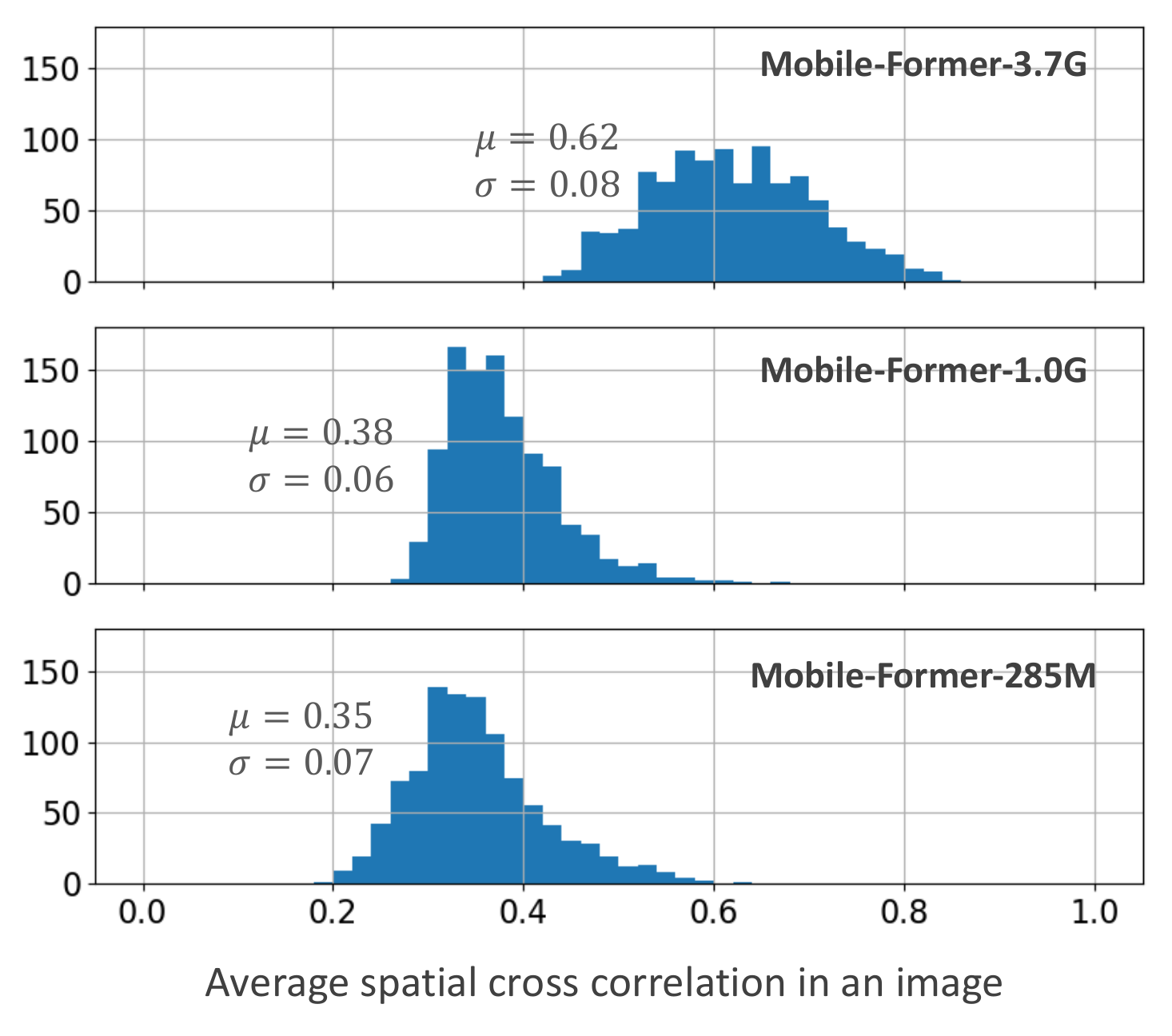}
	\end{center}
	\vspace{-4mm}
	\caption{\textbf{Distribution of spatial correlation in feature maps} over 1000 images. The feature maps are extracted by using MoCo-v2 pre-trained models. Larger models have more spatial correlation than smaller models.}
	\label{fig:moco-corr}
	\vspace{-3mm}
\end{figure}

Below we provide more analysis related to the two \textit{unexpected} behaviors (discussed in \cref{sec:multitask-decoder-probing-results}) in the models pre-trained by MoCo-v2. These two behaviors are: (a) the \texttt{tran-1} probing performance for larger models does not improve when using wider decoders, (b) larger backbones have more degradation in object detection performance. 

We observe a clear difference between large and small models in spatial correlation of output feature maps. The spatial correlation for an image is computed as follows. For a given image with size 224$\times$224, the model outputs a feature map with resolution 14$\times$14 over 196 positions. Following Barlow Twins \cite{zbontar2021barlow}, we use cross-correlation matrix $\mathcal{C}$ computed across spatial positions to represent spatial correlation per image. If all positions are highly correlated, each element in $\mathcal{C}$ is close to $\pm$1. In contrast, if positions are not correlated, $\mathcal{C}$ is close to an identify matrix. For each image, we summarize the spatial correlation as the average of absolute values of off-diagonal elements $\frac{1}{N(N-1)}\sum_i\sum_{j\neq i}|\mathcal{C}_{ij}|$. \cref{fig:moco-corr} plots the histogram of spatial correlation over 1000 validation images in ImageNet. Clearly, the largest model (MF-3.7G) has significantly more spatial correlation than the smaller counterparts (MF-1.0G, MF-285M).

The larger models have stronger spatial correlation because they are more capable to achieve the goal of contrastive learning, i.e. learning common and discriminative features across multiple views. We conjecture this may limit its capability in spatial representation without explicitly modeling spatial relationship like linear prediction in our QB-Heat. As a result, the following decoders in both image classification and object detection have less room to extract more representative features by fusing different spatial positions. And this disadvantage is enlarged when using DETR in object detection, as it heavily relies on spatial representation to regress objects from sparse queries.

Please note that the difference in spatial correlation (between large and small models) is related to, but \textit{not sufficient} to explain the degradation of large models in decoder probing (both classification and detection). We will study it in the future work.

%%%%%%%%%%%%%%%%%%%%%%%%%%%%%%%%%%%%%%%%%%%%%%%%%%%%%%%%%%%%%%
\section{Visualization} \label{apx:vis}

\begin{figure*}[t]
	\begin{center}
		\includegraphics[width=1.0\linewidth]{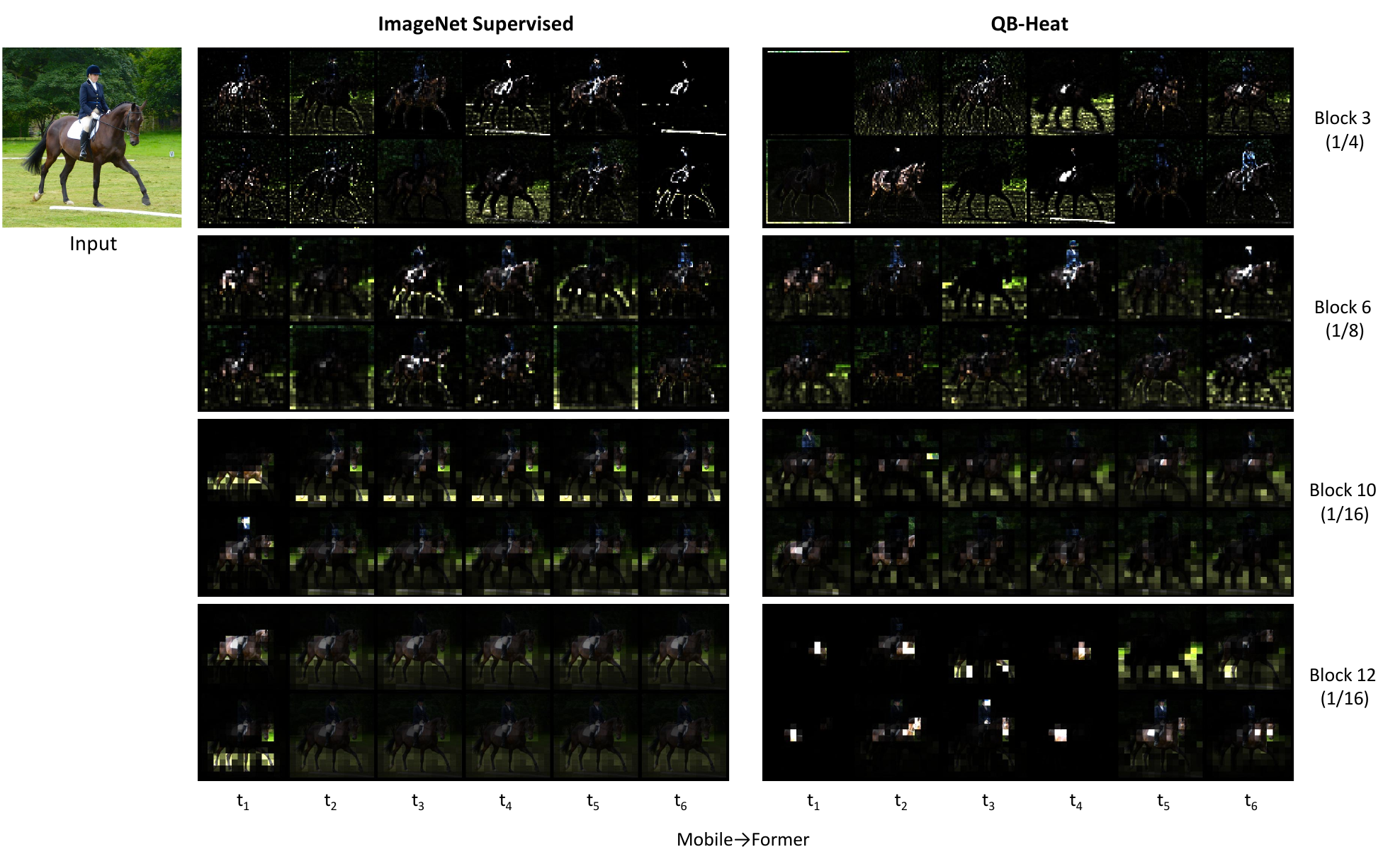}
	\end{center}
	\vspace{-4mm}
	\caption{\textbf{QB-Heat vs. ImageNet supervised pre-training on the cross attention Mobile$\rightarrow$Former}. Mobile-Former-3.7G is used, which includes six tokens (each corresponds to a column). Four blocks at different resolutions are visualized and each has two attention heads visualized in two rows. Attention in \textit{Mobile$\rightarrow$Former} is normalized over pixels, showing the focused region per token. QB-Heat has more diverse cross attentions across tokens (especially at high levels), focusing on different objects (e.g. person, horse, background) and different parts of an object (head, torso, legs of the horse). Best viewed in color.}
	\label{fig:m2f}
\end{figure*}

\begin{figure*}[t]
	\begin{center}
		\includegraphics[width=1.0\linewidth]{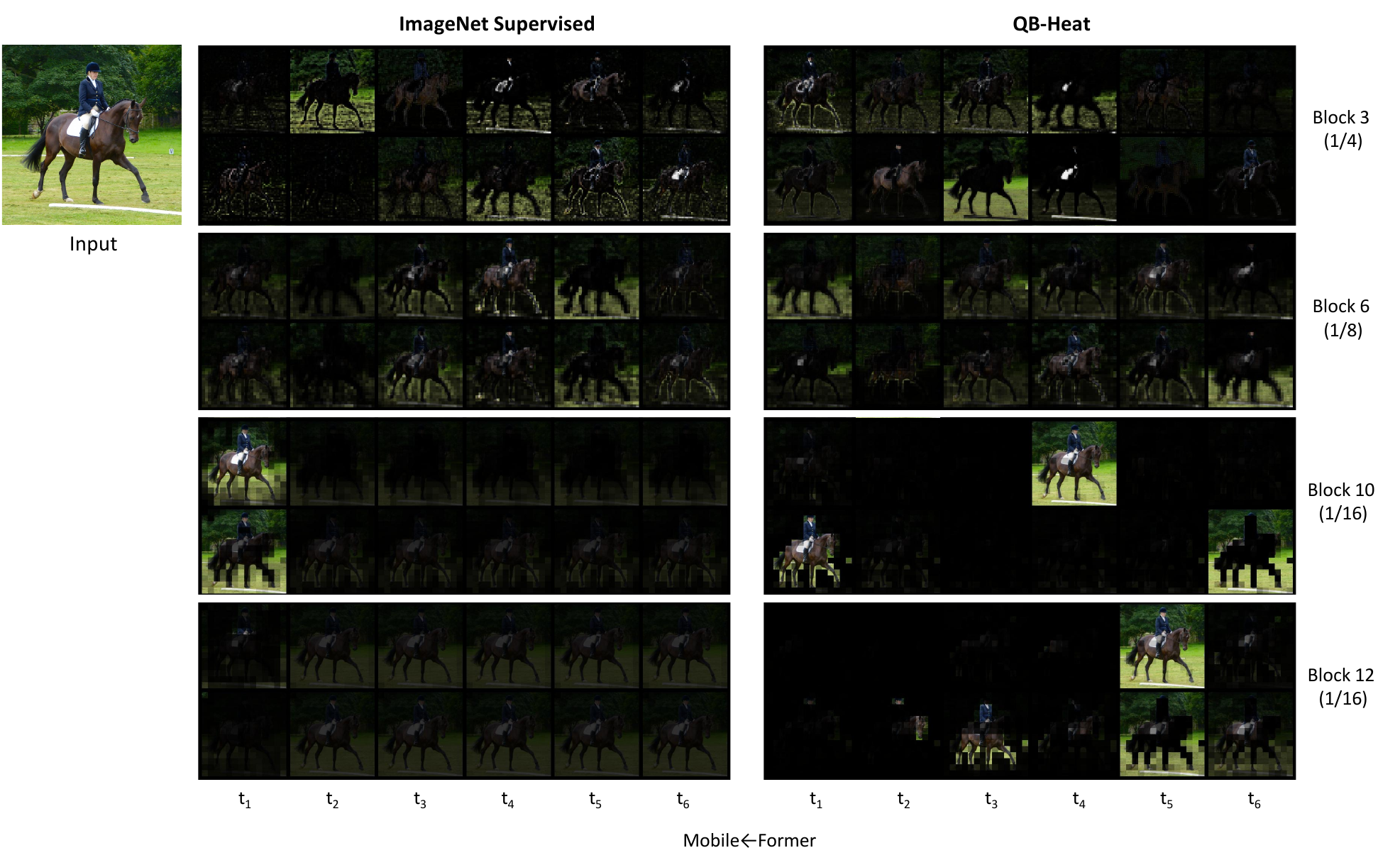}
	\end{center}
	\vspace{-4mm}
	\caption{\textbf{QB-Heat vs. ImageNet supervised pre-training on the cross attention Mobile$\leftarrow$Former}. Mobile-Former-3.7G is used, which includes six tokens (each corresponds to a column). Four blocks at different resolutions are visualized and each has two attention heads visualized in two rows. Attention in \textit{Mobile$\leftarrow$Former} is normalized over tokens showing the contribution of different tokens at each pixel. QB-Heat has more diverse cross attentions across tokens (especially at high levels), segmenting the image into multiple semantic parts (e.g. foreground, background). Best viewed in color.}
	\label{fig:f2m}
\end{figure*}

We also compare our QB-Heat with ImageNet supervised pre-training via visualization of pre-trained models. Following \cite{MobileFormer-2022-cvpr}, we visualize the cross attention on the two-way bridge (i.e. \textit{Mobile$\rightarrow$Former} and \textit{Mobile$\leftarrow$Former}) in \cref{fig:m2f} and \cref{fig:f2m}. Mobile-Former-3.7G is used, which includes six global tokens and eleven Mobile-Former blocks. 

Clearly, QB-Heat has more diverse cross attentions across tokens (especially at high levels). \cref{fig:m2f} shows the cross attention over pixels in \textit{Mobile$\rightarrow$Former}. Compared to the supervised pre-training where tokens share the focus on the most discriminative region (horse torso and legs) at high levels (block 10, 12), QB-Heat has more diverse cross attentions, covering different semantic parts. \cref{fig:f2m} shows another cross attention in \textit{Mobile$\leftarrow$Former} over six tokens for each pixel in the feature-map. QB-Heat also has more diverse cross attentions than the supervised counterpart at high levels, segmenting the image into multiple semantic parts (e.g. foreground, background). This showcases QB-Heat's advantage in learning spatial representation, and explains for its strong performance in multi-task (classification and detection) decoder probing.

\end{document}